\documentclass[twoside,11pt]{article}

\usepackage{blindtext}

\usepackage{jmlr2e}

\usepackage[utf8]{inputenc} 
\usepackage[T1]{fontenc}    
\usepackage{url}            
\usepackage{booktabs}       
\usepackage{amsfonts}       
\usepackage{nicefrac}       
\usepackage{xcolor}         

\usepackage{floatrow}
\usepackage{multirow}
\usepackage{amsmath}
\let\proof\relax

\usepackage{amsthm}
\usepackage{algorithm}
\usepackage{algorithmicx}
\usepackage{algpseudocode}
\usepackage{enumitem}
\usepackage{wrapfig}
\usepackage{caption}
\usepackage{floatrow}
\usepackage{subcaption}
\usepackage{makecell}

\newfloatcommand{capbtabbox}{table}[][\FBwidth]

\usepackage{mathtools}

\newcommand{\ud}{\,\mathrm{d}}

\newcommand{\R}{\mathbb{R}}

\newcommand{\ip}[3]{\left< {#1}, {#2} \right>_{#3}}

\newcommand{\cut}[1]{}
\newcommand{\der}[2]{\frac{\ud {#1}}{\ud {#2}}}
\newcommand{\pder}[2]{\frac{\partial {#1}}{\partial {#2}}}

\DeclareMathOperator{\sinc}{sinc}

\theoremstyle{plain}
\newtheorem{thrm}{Theorem}[section]

\usepackage{lastpage}
\jmlrheading{25}{2024}{1-\pageref{LastPage}}{2/23}{4/24}{23-0137}{Yuxin Sun, Dong Lao, Anthony Yezzi and Ganesh Sundaramoorthi.}
\ShortHeadings{A PDE-based Explanation of Numerical Sensitivities in Training Neural Networks}{Sun, Lao, Yezzi and Sundaramoorthi.}
\firstpageno{1}

\begin{document}

\title{A PDE-based Explanation of Extreme Numerical Sensitivities and Edge of Stability in Training Neural Networks}

\author{\name Yuxin Sun \email syuxin3@gatech.edu \\
       \addr
       Georgia Institute of Technology\\
       Atlanta, GA 30332, USA
       \AND \name Dong Lao
       \email lao@cs.ucla.edu\\
       \addr University of California\\
       Los Angeles, CA 90095, USA
       \AND \name Anthony Yezzi
       \email ayezzi@gatech.edu\\
       \addr  Georgia Institute of Technology\\
       Atlanta, GA 30332, USA
       \AND \name Ganesh Sundaramoorthi
       \email ganesh.sundaramoorthi@rtx.com\\
       \addr Raytheon Technologies\\
       East Hartford, CT 06108, USA
       }

\editor{Francesco Orabona}

\maketitle

\begin{abstract}
We discover restrained numerical instabilities in current training practices of deep networks with stochastic gradient descent (SGD), and its variants. We show numerical error (on the order of the smallest floating point bit and thus the most \emph{extreme} or limiting numerical perturbations induced from floating point arithmetic in training deep nets can be amplified significantly and result in significant test accuracy variance (\emph{sensitivities}), comparable to the test accuracy variance due to stochasticity in SGD. We show how this is likely traced to instabilities of the optimization dynamics that are restrained, i.e., localized over iterations and regions of the weight tensor space. We do this by presenting a theoretical framework using numerical analysis of partial differential equations (PDE), and analyzing the gradient descent PDE of convolutional neural networks (CNNs). We show that it is stable only under certain conditions on the learning rate and weight decay. We show that rather than blowing up when the conditions are violated, the instability can be restrained. We show this is a consequence of the non-linear PDE associated with the gradient descent of the CNN, whose local linearization changes when over-driving the step size of the discretization, resulting in a stabilizing effect. We link restrained instabilities to the recently discovered Edge of Stability (EoS) phenomena, in which the stable step size predicted by classical theory is exceeded while continuing to optimize the loss and still converging. Because restrained instabilities occur at the EoS, our theory provides new insights and predictions about the EoS, in particular, the role of regularization and the dependence on the network complexity.\footnote{Code: \url{https://github.com/sunyx523/surprising-instabilities}}
\end{abstract}


\section{Introduction}

Deep learning is standard practice in a number of application areas including computer vision and natural language processing.  However, the practice has outpaced the theory.  Theoretical advances in deep learning could improve practice, and could also speed the adoption of the technology in e.g., a number of safety-critical application areas.

In this paper\footnote{An earlier version of this manuscript has been published in the Conference of Neural Information Processing Systems, 2022 \citep{sun2022surprising}.}, we discover a phenomenon of \emph{restrained instabilities} in current deep learning training optimization.  Numerical instabilities are present but are restrained so that there is no full global divergence, but these restrained instabilities still cause tiny numerical finite precision errors to amplify, which leads to significant variance in the test accuracy.  We introduce a theoretical framework using tools from numerical partial differential equations (PDE) to provide insight into the phenomena and analytically show how such restrained instabilities can arise as a function of learning rate and weight decay for CNN models. We show that such restrained instabilities arise from over-driving the theoretically predicted learning rate (from PDE stability analysis) of certain non-linear PDEs. Further more, we relate the phenomenon of restrained instabilities to the recently discovered phenomenon of the Edge of Stability (EoS) \citep{DBLP:journals/corr/abs-2103-00065}, in which it is empirically noted that deep learning training through gradient descent often happens at learning rates beyond the stability limit predicted by classical optimization theory, suggesting new theory is needed to understand deep learning optimization. Our theory thus provides insight into EoS from a PDE perspective, and also makes new predictions beyond what is currently known theoretically about EoS.
Our study is a step toward principally choosing learning rates in deep network training and a step toward understanding and constructing deep learning optimization methods from the perspective of numerical conditioning and PDEs. 

In deep learning practice, it is well known that large learning rates can cause divergence and small learning rates can be slow and cause the optimization to be stuck at high training error \citep{bengio2012practical,goodfellow2016deep}. Learning rates and schedules are often chosen heuristically to address this trade-off. On the other hand, in numerical PDEs, there is a mature theory for choosing the learning rate \citep{trefethen1996finite} or step size for discrete schemes for solving PDEs. There are known conditions on the step size (e.g., the Courant-Friedrichs-Lewy (CFL) condition \citep{courant1967partial}) to ensure that the resulting scheme is stable and converges. These methods would thus be natural to study learning rate schemes for training deep networks with SGD, which in the continuum limit is a PDE. However, to the best of our knowledge, numerical PDEs have not been applied to this domain. This might be for two reasons. First, the complicated structure of deep networks with many layers and non-linearities makes it difficult to apply these analytic techniques. Second, the stochastic element of SGD complicates applying these techniques, which primarily do not consider stochastic elements. While we do not yet claim to overcome all these issues, we do show the relevance of PDE analysis in analyzing training of deep networks with SGD, which gives promise for further developing this area.

Our particular contributions are: {\bf 1.}  We found convincing evidence of and thus discovered the presence of restrained numerical instabilities in standard training practices for deep neural networks. {\bf 2.} We show that this results in inherent floating points arithmetic errors having divergence effects on the training optimization that are even comparable to the divergence due to stochastic variability in SGD. {\bf 3.} We present and study the gradient (and accelerated gradient) descent of a simplified CNN model and multi-layer extensions, which allows us to exploit numerical PDE analysis, in order to offer an explanation on how these instabilities might arise. We show in particular that restrained instabilities arise from over-driving the learning rate or step size of certain non-linear PDE beyond their CFL limits. {\bf 4.} We demonstrate that restrained instabilities are likely to occur at the Edge of Stability (EoS), and thus our theory sheds new insight into the EoS, in particular, how this phenomenon might arise in (multi-layer) CNNs.  Our theory also makes new predictions about EoS, in particular, the role of regularization and how such instabilities are more likely to arise in larger multi-layer networks.

\cut{
\footnote{While this first step to study deep networks using more matured tools for the numerical analysis of discretized PDE's is focused on a 1-layer CNN, we show in the appendix how this starting point still yields useful insight into the more complicated dynamics of multi-layer networks, especially when highly regularized, thereby offering strong promise for further developing this theory.}
}

These insights suggest to us that the learning rate should be chosen adaptively over localized space-time regions of the weight tensor space in order to ensure stability while not compromising speed.  While we understand this is a complex task, without a simple solution, our study motivates the need for investigating such challenging strategies.  We note that the subject of training instabilities has been of recent interest to practitioners \citep{chowdhery2022palm,zhang2022opt} in training language models with billions of parameters. Mitigation of these instabilities is desired and no convenient solution exists. Such practical applications motivate the need for a theoretical framework for understanding instabilities in deep networks, which could result in the design of new practical adaptive methods ensuring stability. Note there are a number of heuristically driven methods for adaptive learning rates (e.g., \citep{duchi2011adaptive,zeiler2012adadelta,kingma2014adam,bengio2015rmsprop,luo2018adaptive}), but they do not address numerical stability. In fact, we empirically show training through Adam \citep{kingma2014adam} also exhibits restrained instability.

\section{Related Work}

As stated earlier, we are not aware of related work in applying numerical PDE analysis to study stability and choice of learning rates of deep network training \footnote{In deep learning, ``stability'' often refers to vanishing/exploding gradients, network dynamics, or model robustness. These topics are different from the numerical stability of the training optimization studied in this paper.}. There have been works on studying the convergence of SGD and learning rates (e.g., \citep{latz2021analysis,rakhlin2011making,toulis2016towards}), but they are often based on convexity and Lipschitz assumptions that could be difficult to apply to deep networks.

More broadly, although not directly related to our work, there has been a number of recent works in connecting PDEs with neural networks for various end goals (see \citep{weinan2020towards, karniadakis2021physics, burger2021connections} for detailed surveys). For example, applying neural networks to solve PDEs \citep{sirignano2018dgm,chen2018neural,long2018pde,khoo2021solving,han2018solving}, and interpreting the forward pass of neural networks as approximations to PDEs \citep{ruthotto2020deep, haber2017stable, weinan2017proposal}. Since SGD can be interpreted as a discretization of a PDE, PDEs also play a role in developing optimization schemes for neural networks \citep{chaudhari2018stochastic, chaudhari2018deep, wilson2016lyapunov, osher2018laplacian, lao2020channel, sundaramoorthi2018variational, sun2021accelerated}.

Despite various methods \citep{defazio2014saga,johnson2013accelerating,lei2017non} proposed to reduce the variance of final accuracy induced by the stochastic gradient, we find quite surprisingly that even if all random seeds in optimization are fixed, variance induced by epsilon-small numerical noise (perturbing only the last floating point bit) and hence floating point arithmetic is nevertheless greater or similar to variance across independent trials (see Table \ref{tab:var_relu}).  This observation potentially indicates that numerical stability is a critical but long-ignored factor affecting neural network optimization.

\cut{
This is a manifestation of numerical instability of the discretization of the underlying PDE.
}

There has been work on training deep networks using limited numerical precision (e.g., half precision) \citep{courbariaux2014training,gupta2015deep}. The purpose of these works is to introduce schemes to retain accuracy, which is well known to be reduced significantly when lowering precision to these levels. Different from this literature, we show an unknown fact: errors just in the last bit of floating point representations result in significant divergence. This is to provide evidence of numerical instability resulting from learning rate choices.

Recently, the PDE-based theory for numerical stability introduced in this paper was applied to study and stabilize the training of Neural Implicit Models, i.e., neural networks that represent geometric objects (e.g., surfaces) that are now heavily used in computer vision and computer graphics \cite{yang2024stabilizing}.

{\bf Edge of Stability (EoS)}: Since the publication of our conference version \citep{sun2022surprising}, we discovered a link between restrained instabilities and a recently discovered phenomena called the Edge of Stability (EoS) \citep{DBLP:journals/corr/abs-2103-00065}, and thus our theory sheds insight into EoS. In \citep{DBLP:journals/corr/abs-2103-00065}, it is reported empirically that training of deep networks through gradient descent (GD) does not obey predictions from classical optimization theory. Classical theory predicts that optimization is stable when the sharpness (maximum eigenvalue of the Hessian) is less than 2/lr (where lr is the learning rate). However, GD of deep networks increases its sharpness to above 2/lr and then oscillates slightly above this value. This suggests a new theory is needed and hence several works aim to provide a theory \citep{DBLP:journals/corr/abs-2002-09572,https://doi.org/10.48550/arxiv.2204.11326,https://doi.org/10.48550/arxiv.2204.01050,https://doi.org/10.48550/arxiv.2209.15594,https://doi.org/10.48550/arxiv.2207.12678}. Note that \cite{NEURIPS2018_6651526b} first discovered that GD converges to the local minima with sharpness close to 2/lr.

\citep{https://doi.org/10.48550/arxiv.2204.01050} suggests that this could arise from (non-linear) activations, and motivate this by looking at a quadratic function followed by a tanh. It is shown that GD can unstably converge to a stable set. Further, a relation between loss minimization and oscillations in GD is given to explain the overall decrease of the loss. In \citep{arora2022understanding}, the EoS is attributed to an implicit bias of GD in general. Analysis is provided for GD of the square root of loss and normalized GD; it is shown that there are two phases, one that tracks gradient flow (and sharpness increases to 2/lr) and the second that oscillates around the minimum loss manifold. The analysis is limited around the minimum loss manifold, and analyzing EoS for GD on the original loss is future work and would probably require the use of specific properties of the loss the authors state. \citep{https://doi.org/10.48550/arxiv.2209.15594}, analyzes GD of the loss itself and attributes EoS to an implicit bias of GD called self-stabilization.  They show that as GD blows up due to the sharpness exceeding 2/lr, self-stabilization decreases the sharpness, which is shown through a third-order Taylor approximation of the loss that relates to the sharpness gradient. 

Despite attributing EoS to general properties of GD \citep{arora2022understanding,https://doi.org/10.48550/arxiv.2209.15594}, understanding what specifics of training neural networks contributes to EoS is needed. To this end, there has been work to understand EoS on simple examples exhibiting properties of neural networks. For example, in \citep{chen2022gradient}, a 2-layer single neuron network is analyzed. In \citep{zhu2022understanding}, the authors analyze a 4-layer scalar network (without non-linearities).  In contrast to existing work on EoS, we use numerical PDE tools to analyze the stability of GD (and Nesterov GD) on arbitrary number of layer multi-layer model of CNNs. We introduce a concept of \emph{restrained instabilities} and show it arises from violating the strict CFL condition of the discretization of certain non-linear PDE. Instabilities that arise locally change the linearization of the PDE, which has a stabilizing effect, causing oscillations. We show restrained instabilities are at EoS. Our analysis also extends to arbitrarily large multi-layer CNNs and shows how such restrained instabilities arise locally at a single layer and can propagate through the network.

\section{Background: PDEs, Discretization, and Numerical Stability}
\label{sec:background}
We start by providing background on analyzing the stability of PDE discretizations, and introducing the concepts of (in)stability and restrained instability. This methodology will be applied to neural networks in later sections as a PDE is the continuum limit of SGD.

\subsection{PDE Stability Analysis Illustrated on the Heat Equation}

We illustrate concepts using a basic PDE, the heat equation, which models a diffusion process:
\begin{equation}
    \partial_t u(t,x) = \kappa \partial_{xx} u(t,x), \quad u(0,x) = u_0(x),
    \label{eq:heat1D}
\end{equation}
where $u : [0,T] \times \R \to \R$, $\partial_t u$ denotes the partial with respect to time $t$, $\partial_{xx} u$ is the second derivative with respect to the spatial dimension $x$, $\kappa>0$ is the diffusivity coefficient, and $u_0 : \R \to \R$ is the initial condition. To solve this numerically, one discretizes the equation, using finite difference approximations. A standard discretization is through $\partial_{xx} u(t,x) = \frac{u(t,x+\Delta x)-2u(t,x)+u(t,x-\Delta x)}{(\Delta x)^2} + O((\Delta x)^2)$ and $\partial_t u(t,x)=\frac{u(t+\Delta t,x)-u(t,x)}{\Delta t} + O(\Delta t)$, where $\Delta x$ is the spatial increment and $\Delta t$ is the step size, which gives the following update scheme:
\begin{equation}
    u^{n+1}(x) = u^{n}(x) + \frac{\kappa\Delta t}{(\Delta x)^2} \cdot [u^n(x+\Delta x)-2u^n(x)+u^n(x-\Delta x)] + \epsilon^n(x),
    \label{eq:fde}
\end{equation}
where $n$ denotes the iteration number, $u^n(x)$ is the approximation of $u$ at $t = n\Delta t$, and $\epsilon$ denotes the error of the approximation (i.e., due to discretization and finite precision arithmetic).

A key question is whether $u^n$ remains bounded as $n\to \infty$, i.e., \emph{numerically stable}. It is typically easier to understand stability in the frequency domain, which is referred to as Von Neumann analysis \citep{trefethen1996finite,richtmyer1994difference}.
To do this, we recall the Discrete Fourier Transform (DFT). Let the spatial sampling of a function, $u$, be $\{u(x_k)\}_{k=0}^{K-1}$ where $x_k=k\Delta x$ for $k\!=\!0,\ldots,K\!-\!1$ are the discrete samples of the continuum variable $x$, and $K$ is the number of samples. Let $\omega_m=m\frac{2\pi}{K}$ for $m\!=\!0,\ldots,K\!-\!1$ represent discrete frequencies, then the DFT, denoted $\hat u$, is given by
$$\hat u(\omega_m) = \sum_{k=0}^{K-1} u(x_k)\,e^{-i\omega_m k},$$
where $i$ is the imaginary unit. Computing the spatial DFT of \eqref{eq:fde} yields:
\begin{equation}
    \hat u^{n+1}(\omega_m) = A(\omega_m)\hat u^n(\omega_m) + \hat \epsilon^n(\omega_m),\quad \mbox{or}\quad 
    \hat u^n(\omega_m) = A^n(\omega_m)\hat u_o(\omega_m) + \sum_{i=0}^{n-1} A^i(\omega_m)\hat\epsilon^{n-i}(\omega_m),
\end{equation}
 and $$A(\omega) = 1-\frac{2\kappa\Delta t}{(\Delta x)^2} [1-\cos(\omega\Delta x)]$$ is the amplifier function. Note that $u^n$ is stable if and only if $|A(\omega_m)|<1$ for all $\omega_m\in\{\omega_0,\ldots,\omega_{K-1}\}$. Notice that if $|A(\omega)|<1$ for all $\omega$ (a slightly stronger but much easier condition to analyze), then $u^n$ converges and errors $\epsilon^n$ are attenuated over iterations. If $|A(\omega_m)|\geq 1$ for any $\omega_m\in\{\omega_0,\ldots,\omega_{K-1}\}$, then $u^n$ diverges and the errors $\epsilon^n$ are amplified at these frequencies. The case, $|A(\omega)|<1$ for all $\omega$, implies the conditions $\Delta t > 0$ and 
\begin{equation}
\Delta t < \frac {(\Delta x)^2}{2\kappa}; \tag{\mbox{CFL Condition for Heat PDE}}
\end{equation}
this is known as the CFL condition. Notice the restriction on the time step (learning rate) for numerical stability. If the discretization error of the operator on the right hand side of the PDE is less than $O(\Delta x)$, stability also implies convergence to the PDE solution as $\Delta t \to 0$ \citep{trefethen1996finite,richtmyer1994difference}.

\begin{figure}
\begin{minipage}[h]{0.5\linewidth}
\centering
\includegraphics[width=0.8\columnwidth]{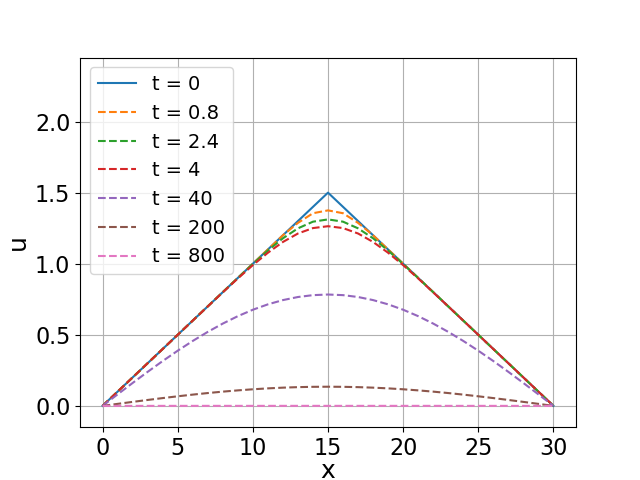}
\label{fig:heat}
\end{minipage}%
\begin{minipage}[h]{0.5\linewidth}
\centering
\includegraphics[width=0.8\columnwidth]{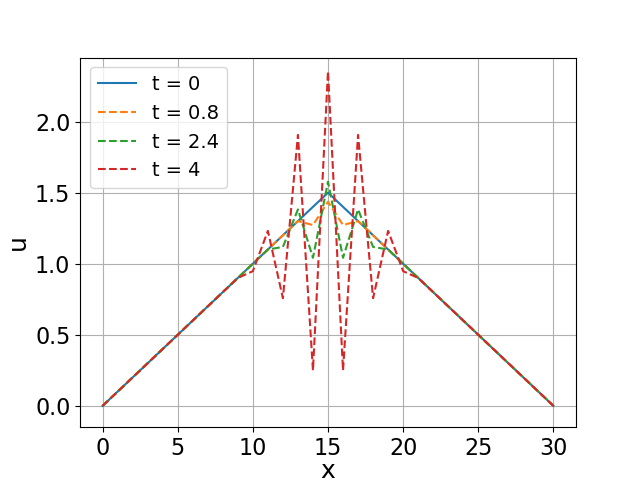}
\caption{Illustration of instability in discretizing the heat equation. The initial condition $u_0$ is a triangle (blue), with boundary conditions $u(0)=u(30)=0$. {\bf [Left]}: When the CFL condition is met, i.e., $\kappa \Delta t/ (\Delta x)^2 = 0.4 < \frac 1 2$, the method is stable and approximates the solution of the PDE. Note the true steady state is 0, which matches the plot. {\bf [Right]}: When the CFL condition is not met, i.e., $\kappa \Delta t/(\Delta x)^2 = 0.8 > \frac 1 2$, small numerical errors are amplified and the scheme diverges.}
\label{fig:heat_discrete}
\end{minipage}%
\end{figure}

See Figure~\ref{fig:heat_discrete} for simulation of the discretization scheme for the heat PDE. The instability starts locally but quickly spreads and causes blowup of the solution globally. For standard optimization of CNNs, we will show rather that the instability is \emph{restrained} (localized in time and space), which nevertheless causes divergence.  

\subsection{Illustration of Restrained Instabilities Through the Beltrami PDE}

We illustrate a new concept of restrained instabilities. To the best of our knowledge, we
are not aware that this phenomenon has been documented in numerical PDE or the ML community. We show restrained instabilities are a non-linear PDE phenomenon, and as such, we illustrate it with the Beltrami reaction-diffusion PDE, which is a non-linear PDE.
The Beltrami reaction-diffusion PDE and the loss that it optimizes is
\begin{equation}
   \partial_t u =\partial_x\left(\frac{\partial_x u}{\sqrt{(\partial_x u)^2+\delta^2}}\right)
                +\lambda\,(I-u), \quad 
    L(u) = \int_{\R} \left[ \frac{\lambda}{2} (I-u)^2 + \sqrt{ (\partial_x u)^2+\delta^2 }\right]  \ud x
   \label{eq:beltrami}
\end{equation}
where $\delta>0$ is a small constant that ensures differentability of the loss, $I : \mathbb{R}\to\mathbb{R}$ is a given function, and $\lambda>0$ is the data fidelity weight. Note that this PDE trades-off between fitting the function $I$ and being smooth while preserving discontinuities; this has been used in image denoising \citep{rudin1992nonlinear}. To analyze stability, we cannot directly use Von-Neumann analysis since the PDE is non-linear, however, we may study the linearization around a function and apply the analysis.  The non-linear PDE is stable if and only if the linearization is stable.

A linearization is obtained by treating $\kappa$ as locally constant and Beltrami PDE is then similar to the heat PDE but with the extra fidelity term. Using similar analysis as in the previous section gives that
\begin{equation}
\Delta t < \frac{1}{\lambda + 2\kappa/(\Delta x^2)}, \quad 
\kappa = [(\partial_x u)^2+\delta^2]^{-1/2}
\tag{CFL Condition for Beltrami PDE}
\end{equation}
Note that the step size is dependent on the local structure of the gradient of $u$. If we want strict stability then we must choose 
$
\Delta t <  \frac{\Delta x^2}{2}\delta,
$
(in the case $\lambda$ is small) the worst-case step size overall $x$, which is excruciatingly small. Consider the case when we exceed this strict stability condition. The step size chosen would exceed the stable step size in neighborhoods with small spatial gradients, and thus the evolution is unstable. However, as the instability grows, in these neighborhoods, the gradient of $u$ grows\footnote{Note that due to discretization and finite precision errors, high-frequency components are present (though small), an instability exponentially grows such components and thus the spatial gradient of the function grows.} (changing the local linearization) and thus relaxes (increases) the stable step size. This in turn restrains the instability and the regularization term smooths $u$, which in turn reduces the spatial gradient (again changing the local linearization and restricting the stable step size to its previous lower value), and the process can continue to oscillate. Thus, neighborhoods could go back and forth between unstable and stable.  These oscillations persist but are nevertheless never able to grow without bounds. The larger the time step beyond the strictly stable step size, the more the instabilities are allowed to grow before being restrained. Note as the chosen step size exceeds the stable step size for every point, the process is fully unstable, and leads to a full blowup similar to when the CFL condition is exceeded in the heat equation. In later sections, we provide evidence that such restrained instabilities occur in training neural networks.

Figure~\ref{fig:restrained_beltrami_pde} shows the results of optimizing the loss associated with the Beltrami PDE for step sizes exceeding the maximum stable step size, producing restrained instabilities. For reference, the result for the stable step size is also shown.
We see the restrained instability phenomenon for time steps that exceed the strict stability condition by a factor of 10 and 100. In the former case, we converge to a similar answer as in the stable case, but with very low amplitude background oscillations, while in the latter case, the persistent oscillations have a larger amplitude yet are still restrained.

\cut{
\begin{figure}[h]
\begin{minipage}[h]{0.5\linewidth}
\centering
\includegraphics[width=\columnwidth,clip,trim={80 15 80 80}]{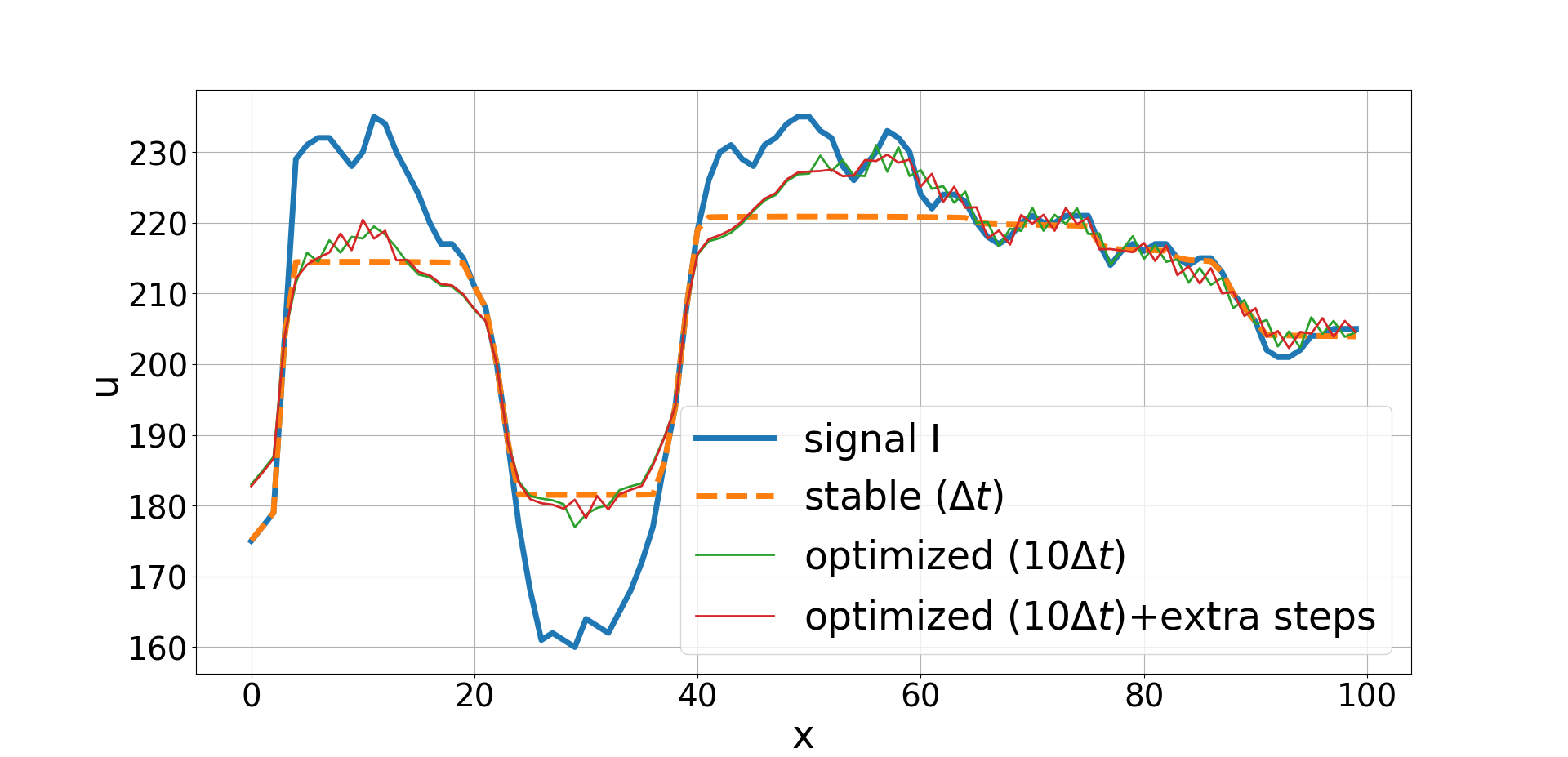}
\end{minipage}%
\begin{minipage}[h]{0.5\linewidth}
\centering
\includegraphics[width=\columnwidth,clip,trim={80 15 80 80}]{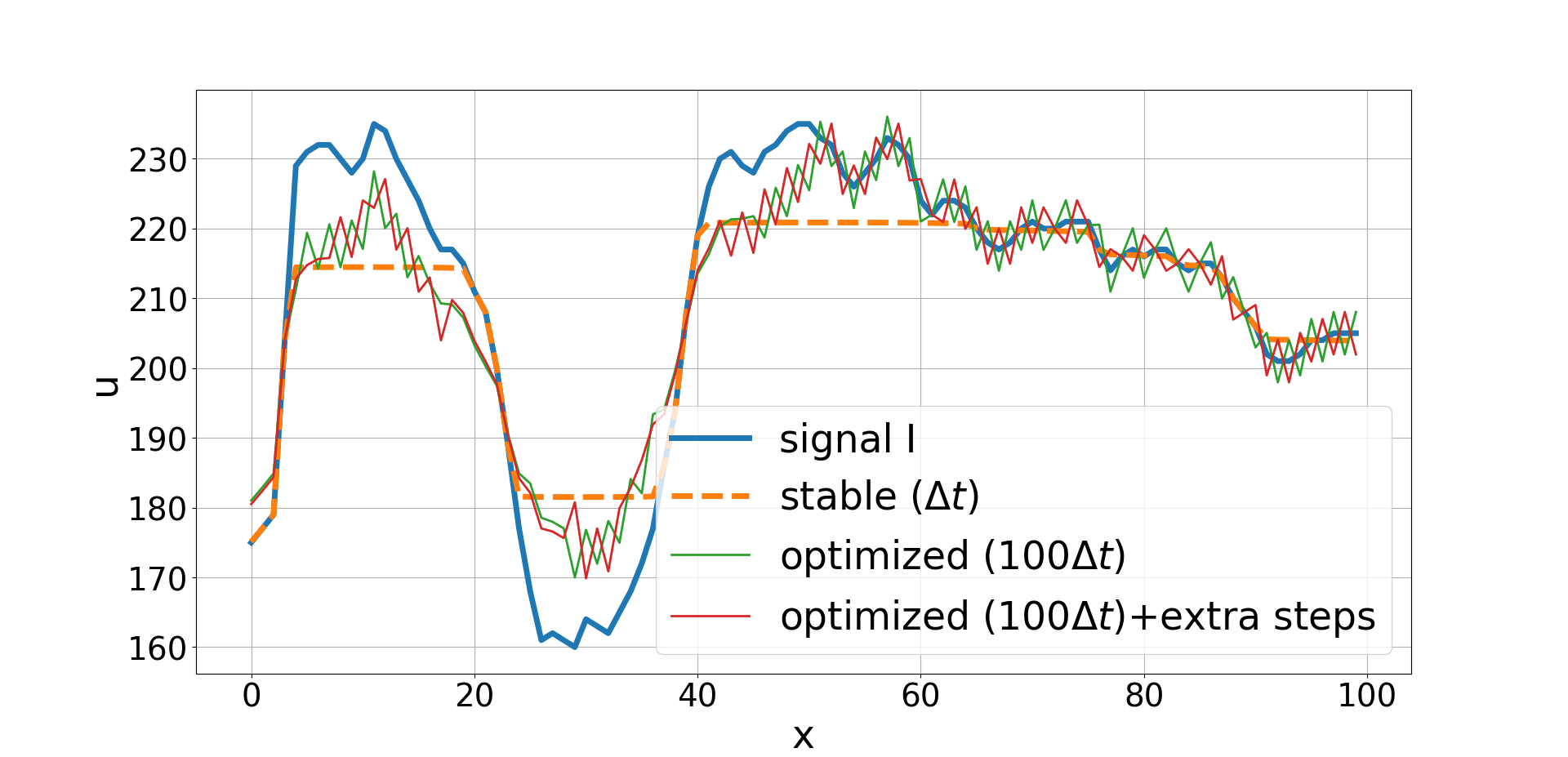}
\caption{  {\bf Demonstration of restrained instability}
  related to ``moderately aggressive'' choices of step size in the
  discretized Beltrami PDE. While temporary instabilities continuously develop locally throughout the optimization process, the oscillations they generate
  change the local linearization which in turn has a local stabilizing effect
  (temporarily) that restrains the instability from exploding. As such, we
  can optimize, despite these restrained instabilities, to obtain a candidate
  minimizer which contains a level of effective ``background-noise'' arising
  from the restrained instabilities when exceeding the maximum stable steps size by a factor
  of 10 (left) and 100 (right). Both converge in a stochastic sense but with
  a degraded signal-to-noise ratio for the larger step size. Continued
  optimization steps beyond this effective convergence (green curves) simply change the
  noise pattern (red curves).}
\label{fig:restrained_beltrami_pde}
\end{minipage}%
\end{figure}
}

\begin{figure}[h]
\begin{minipage}[h]{0.33\linewidth}
\centering
{\small restrained small step size}\\
\includegraphics[width=\columnwidth,clip,trim={85 10 90 70}]{fig/beltrami/restrained10-1D}
\end{minipage}%
\begin{minipage}[h]{0.33\linewidth}
\centering
{\small restrained large step size}\\
\includegraphics[width=\columnwidth,clip,trim={85 10 90 70}]{fig/beltrami/restrained100-1D}
\end{minipage}%
\begin{minipage}[h]{0.33\linewidth}
\centering
{\small unstable}\\
\includegraphics[width=\columnwidth,clip,trim={40 10 100 70}]{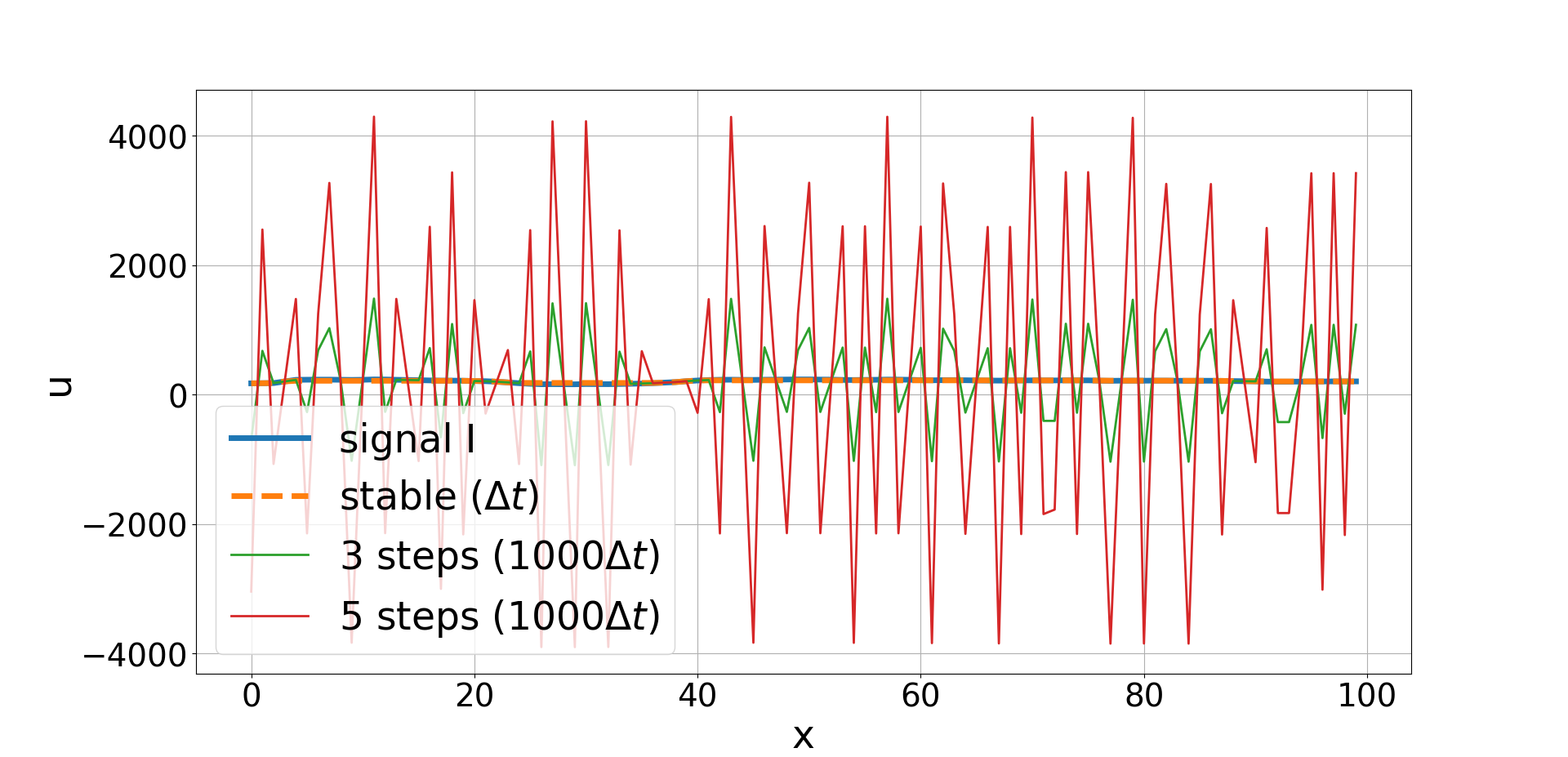}
\end{minipage}%
\caption{  {\bf Demonstration of restrained instability}
  related to ``moderately aggressive'' choices of step size in the
  discretized Beltrami PDE. While temporary instabilities continuously develop locally throughout the optimization process, the oscillations they generate
  change the local linearization which in turn has a local stabilizing effect
  (temporarily) that restrains the instability from exploding. As such, we
  can optimize, despite these restrained instabilities, to obtain a candidate
  minimizer which contains a level of effective ``background-noise'' arising
  from the restrained instabilities when exceeding the maximum stable steps size by a factor
  of 10 (left) and 100 (middle). Both converge in a stochastic sense but with
  a degraded signal-to-noise ratio for the larger step size. Continued
  optimization steps beyond this effective convergence (green curves) simply change the
  noise pattern (red curves). If the stable step size is exceeded too much (e.g., by a factor of 1000 on the right plot), the process becomes unstable without restraint and does not converge even in a stochastic sense.}
\label{fig:restrained_beltrami_pde}
\end{figure}

The previous discussion introduces restrained instabilities at a conceptual level and explains how they might arise, which is the intent in this paper. A formalized definition of a restrained instability could be similar to the analysis we will provide Section~\ref{subsec:restrained_instabilities}: Suppose space-time is divided into different regimes (e.g., different linear regimes) of a PDE where in some of these regimes the linearized PDE is unstable, and some stable. A restrained instability might be defined to occur when the path of the PDE continues to pass back and forth indefinitely through stable and unstable regimes (as classified by the linearization of the system within such regimes), thereby never allowing unstable effects to grow indefinitely nor allowing them to attenuate indefinitely as the system hovers around the shared boundary of a stable/unstable regime. Despite this formalization, even a finite number of transitions through stable and unstable regimes that eventually converges within a stable regime will still benefit from the analysis that we provide in this paper.

\section{Empirical Evidence of Instability in Training Deep Networks}
\label{sec:floatingpoint_error}

We discover and provide empirical evidence of restrained instabilities in current deep learning training practice. To this end, we show that optimization paths in current practice of training convolutional neural networks (CNNs) can diverge significantly due to the smallest errors from finite precision arithmetic. We show that the divergence can be eliminated with learning rate choice.

\subsection{Perturbing SGD by the Smallest Floating Point Perturbations}
We introduce a modified version of SGD that introduces perturbations of the original SGD update on the order of the smallest possible machine representable perturbation, modeling noise on the order of error due to finite precision floating point arithmetic:
\begin{align*}
    \theta_{t+1} &= \theta_t - \eta g_t \tag{\mbox{SGD}} \\
    \theta_{t+1} &= \theta_t - (\eta/k) \times (kg_t), \tag{\mbox{PERTURBED SGD}}
\end{align*}
where $\eta > 0$ is the learning rate, $g_t$ is the stochastic gradient or the momentum vector if momentum is used, and $k>1$ is an odd integer. Notice that mathematically, both updates are exactly equivalent (division by $k$ cancels the multiplication by $k$), however considering floating point arithmetic, these may not be equivalent.  A floating point number is represented as 
\begin{equation}
    \mbox{significand} \times \mbox{base}^{\text{exponent}} \quad 
    (s\,\,\text{bits significand}, e\,\,\text{bits exponent}) 
\end{equation}
where typically base 2 (binary) is used. Thus, the modified update introduces a potential perturbation in the last significant bit of $\eta g_t$, hence a relative difference of the right hand sides on the order of $2^{-s}$ according to IEEE standards on floating point computation. In other words, $\mbox{RelativeError}(x,y) := |x-y|/|x| < 2^{-(s-1)}$ where $x=\mbox{fl}(\eta g_t)$, $y=\mbox{fl} (\eta/k) \times \mbox{fl}(kg_t) $, and $\mbox{fl}(x)$ is the floating point representation of $x$. Note that this is the same order of relative error between $\eta g_t$ and $\mbox{fl}(\eta g_t)$, the inherent machine error due to floating point representation.  The perturbation induces the \emph{smallest} possible perturbation of $\eta g_t$ representable by the machine. Thus, perturbed SGD is a way to show the effects of machine error. Notice that when $k$ is a power of two, as floating point numbers are typically stored in binary form, the division just subtracts the exponent and the multiplication adds to the exponent so both updates are equivalent and no perturbation is introduced. However, choosing $k$ odd can create changes at the last significant bit, with each choice of $k$ yielding a different change.

\subsection{Perturbed SGD on CNNs and Demonstration of Restrained Instabilities} \label{sec:expts_deep_net_instability}
In SGD, there are several sources of randomness, i.e., random initialization, random shuffled batches/selection, random data augmentation, and there are also sources of non-determinism in implementations of the deep learning frameworks.  To ensure the variance merely originated from the floating point perturbation, in the following experiments, we eliminate all randomness by fixing the initialization, the random seed for batch selection, and the non-determinism flag. Appendix~\ref{app:seeds_fixed} empirically validates that all seeds are fixed.
We use Pytorch using default 32-bit floating point precision (similar conclusions for 64-bit hold).

\cut{such as in PyTorch, there is non-determinism in the choice of convolution algorithm and other operators.
in the deep learning framework (e.g., torch.backends.cudnn.deterministic in PyTorch)
}

{\bf Perturbed SGD with ReLU}: Our first experiment uses the ResNet-56 architecture \citep{He_2016_CVPR}, which we train on CIFAR-10 \citep{krizhevsky2009learning} using perturbed SGD. We use the standard parameters for training this network \citep{He_2016_CVPR}: {\tt lr=0.1, batch size = 128, weight decay = 5e-4, momentum=0.9}. The standard step decay learning rate schedule is used: the learning rate is divided by 10 every 40 epochs for a total of 200 epochs. We report the final test accuracies for 6 different values of $k$, including $k=1$ (standard SGD), and compute the standard deviation of the accuracies (STD). Table~\ref{tab:var_relu} (left) shows the results. This results in test accuracy variability over different $k$. We repeat this experiment over different seeds for the stochastic batch selection (shown in the next columns) fixing the initialization. The standard deviation for each seed over $k$ is on average $0.16\%$. To illustrate the significance of this value, we compare to the variability due to batch selection on the right of Table~\ref{tab:var_relu}. We empirically estimate the relative fluctuation of the gradient estimate from batch selection,
which is $26.72$ \footnote{To compute the gradient fluctuation with respect to batches, we sample $B$ batches and compute $x$ to be the average batch gradient over the $B$ batches. We then randomly sample a batch and let $y$ be the batch gradient. The relative gradient fluctuation at an epoch is $\frac{1}{M} \sum_i \frac{|x_i-y_i|}{|x_i|}$ where $M$ is the number of parameters and $x_i$ is the $i^{\text{th}}$ component. We then take the median over epochs as the relative gradient fluctuation.}. This is large compared to $2^{-23}$ for the floating point perturbation, yet this smallest fluctuation that is machine representable (and models finite precision error) results in about the same accuracy variance as batch selection. Thus, finite precision error, inherent in the system, surprisingly results in about the same test accuracy variance as stochastic batch selection.

\cut{
The test accuracy variation due to floating point arithmetic is about the same as the variance due to stochastic batch selection. {\color{red} quantify stochastic gradient variance and compare it to float error}
}

\cut{
\begin{tabular}{cccccccc}
	\toprule
    Seed & 1 & 2 & 3 & 4 & 5 & 6 & std\\
    \midrule
     lr*gd& 93.36 & 93.40 & 93.10 & 93.14 & 93.34 & 93.33 & 0.13\\
	\midrule
	 lr/3*gd*3 & 93.49 & 93.37 & 93.08 & 93.68 & 93.16 & 93.12\\
	 \midrule
	 lr/5*gd*5 & 93.64 & 93.22 & 93.39 & 93.17 & 93.26 & 93.42\\
	 \midrule
	 lr/7*gd*7 & 93.36 & 93.31 & 93.12 & 93.23 & 93.14 & 93.28\\
	 \midrule
	 lr/9*gd*9 & 93.87 & 93.55 & 93.08 & 93.35 & 93.42 & 93.41\\
	 \midrule
	 lr/11*gd*11 & 92.99 & 93.31 & 93.49 & 93.48 & 93.14 & 93.56\\
	 \midrule
	 std & 0.27 & 0.10 & 0.16 & 0.19 & 0.10 & 0.13 & 0.16\\

	\bottomrule
\end{tabular}
}

\begin{table}[h]
\center
\tiny
\begin{sc}
\begin{tabular}{cccccccc}
	\toprule
    Seed & 1 & 2 & 3 & 4 & 5 & 6 & STD \\
    \midrule
     $k=1$ & 93.36 & 93.40 & 93.10 & 93.14 & 93.34 & 93.33 & 0.11 \\
	\midrule
	 $k=3$ & 93.49 & 93.37 & 93.08 & 93.68 & 93.16 & 93.12 & 0.22\\
	\midrule
    $k=5$ & 93.64 & 93.22 & 93.39 & 93.17 & 93.26 & 93.42 & 0.16\\
	\midrule
	$k=7$ & 93.36 & 93.31 & 93.12 & 93.23 & 93.14 & 93.28 & 0.09\\
	\midrule
	$k=9$ & 93.87 & 93.55 & 93.08 & 93.35 & 93.42 & 93.41 & 0.24\\
	\midrule
	$k=11$ & 92.99 & 93.31 & 93.49 & 93.48 & 93.14 & 93.56 & 0.21\\
	\midrule
	std & 0.27 & 0.10 & 0.16 & 0.19 & 0.10 & 0.13 & \\
    \bottomrule
\end{tabular}\hspace{0.1cm}
\begin{tabular}{ccc}
    \toprule
     Perturbation & Relative & Test Acc \\
     Method & Grad. Error & Std \\
     \midrule
     Floating Pt (32 bit) & $ 2^{-23}$ & 0.16 \\
     \midrule
     Floating Pt (64 bit) & $ 2^{-52}$  & 0.15 \\
     \midrule
     Batch Selection & 26.72  & 0.17\\
     \bottomrule
\end{tabular}
\end{sc}
\caption{{\bf [Left]}: Final test accuracy over different seeds (batch selections) and different floating point perturbations (rows) for Resnet56 trained on CIFAR-10. STD is the standard deviation. {\bf [Right]}: Comparison of errors induced in the gradient and resulting test accuracy STD for floating point perturbation vs batch selection. Floating point error perturbs the gradient by an amount of 8 orders of magnitude (in the case of 32 bit representation) smaller than batch selection yet surprisingly yields nearly the same test accuracy variation! Note increasing floating point precision (to 64 bit) does not mitigate the issue. }
\label{tab:var_relu}
\end{table}

\cut{
\begin{wrapfigure}[12]{r}{0.4\textwidth}
\vskip -0.3in
\includegraphics[width=0.9\columnwidth]{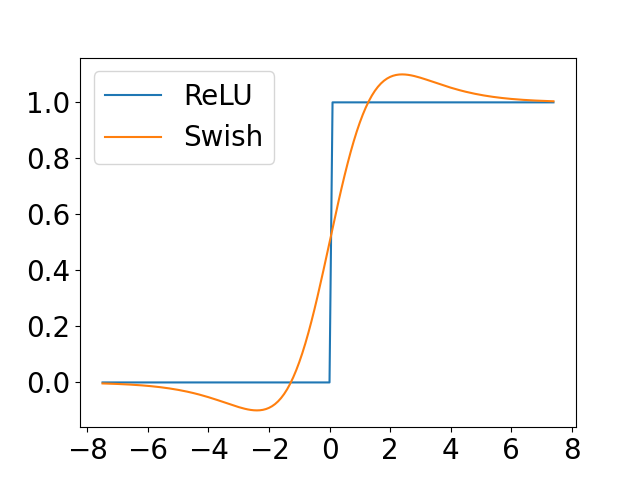}
\caption{Activation Gradients}
\label{fig:act_der}
\end{wrapfigure}
}

{\bf Perturbed SGD with Swish}: We first conjectured that the variance due to floating point errors could be caused by the ReLU activation, the activation standard in ResNets. This is because the gradient of the ReLU is discontinuous at zero. Hence small positive values (e.g., 1e-6) would give a gradient of 1 and small negative values (e.g., -1e-6) would give a gradient of zero. Hence the ReLU gradient is sensitive to numerical fluctuation around zero. Note common deep learning implementations choose a value at zero, among 0, 0.5 or 1; but this has the same sensitivity. Therefore, we adopted the Swish activation function (proposed in \citep{ramachandran2017searching}), $\mbox{Swish}(x) := x/[1+\exp{(-\beta x)}]$ ($\beta>0$, as $\beta\to \infty$ Swish approaches ReLU), which is is an approximation of the ReLU that has a continuous gradient at zero. Recent work has shown the Swish function improves accuracy \citep{ramachandran2017searching} as well. Thus, we repeated the previous experiment (under the same settings) replacing the ReLU with the Swish activation. Results are shown in Appendix~\ref{app:extra_tables}, and is similar to the previous experiment with ReLU: the standard deviation in test accuracy due to floating point perturbation ($93.72 \pm 0.15$) is significant and is comparable (even exceeds) the variance due to stochastic batch selection ($0.09$).

\cut{
\begin{wrapfigure}{r}{0.35\textwidth}
\vskip -0.3in
\includegraphics[width=0.9\columnwidth]{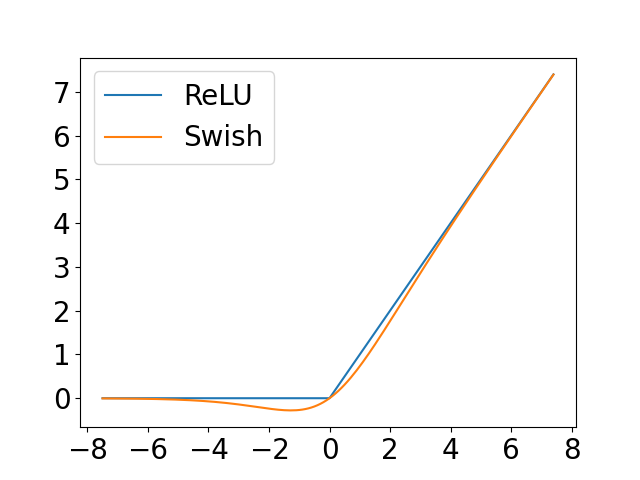}
\caption{Activation Functions}
\label{fig:act}
\end{wrapfigure}
}

{\bf Other Architectures/Datasets/Optimization Algorithms/Data Types}: To verify that this phenomenon is not specific to the choice of network architecture or dataset, we repeated the same experiment for a different network (VGG16 \citep{simonyan2014very}) and a different dataset (Fashion-MNIST \citep{xiao2017fashion}). We also switch the optimization algorithm from SGD to ADAM \citep{kingma2014adam}. In addition, we increase the precision of the floating point representation from 32 bit to 64 bit, and the results indicate the error is independent of the floating point precision. We summarize the results in the Appendix~\ref{app:extra_tables}, which confirms that the phenomena are still present across different networks, datasets, optimization algorithms, and data-types.

\begin{figure}[h]
\begin{minipage}[h]{0.5\linewidth}
\centering
\includegraphics[width=0.8\columnwidth]{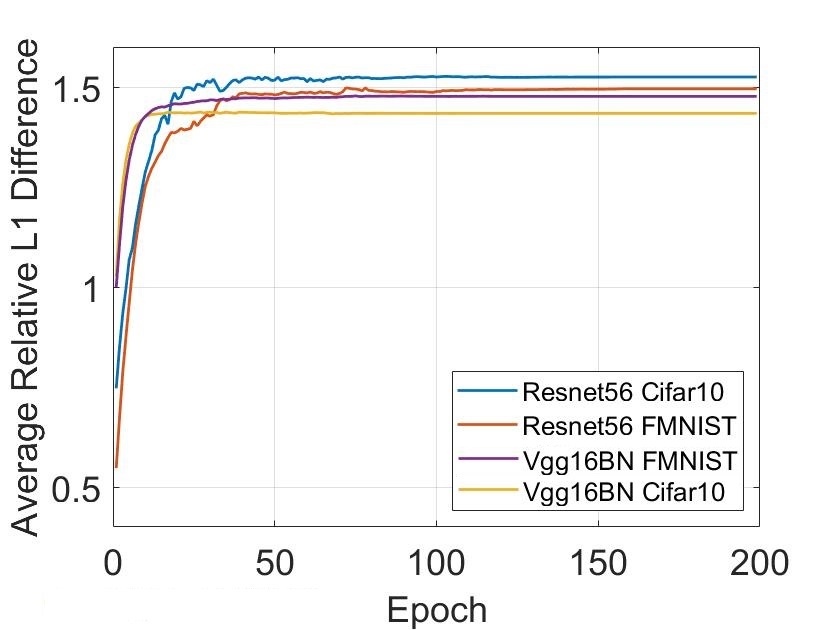}
\label{fig:epoch}
\end{minipage}%
\begin{minipage}[h]{0.5\linewidth}
\centering
\includegraphics[width=0.8\columnwidth]{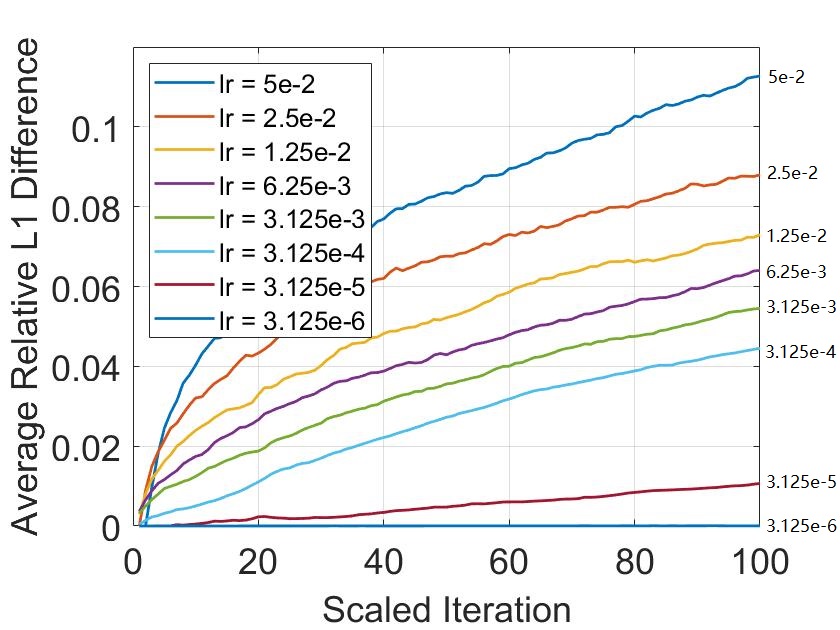}
\caption{{\bf [Left]}: Relative L1 difference in weights across epochs for SGD ($k=1$) and modified SGD ($k=3$). A decayed learning rate schedule is used. The errors quickly build up, but then are contained at higher epochs. The initial build-up of errors is enough to result in different test accuracy. This gives evidence of restrained instability. {\bf [Right]}: Relative L1 difference in weights over initial iterations for SGD ($k=1$) and modified SGD ($k=3$) with various fixed learning rates. Lower than some minimum learning rate, the floating point perturbation gets attenuated and higher than that value errors are amplified, suggesting an instability.}
\label{fig:iteration}
\end{minipage}%
\end{figure}

{\bf Divergence in Optimization Paths}: We plot the difference in network weights (using the measure described next) between the original SGD weights, $\theta_i$ ($i$ is the index for location in the vector), and the perturbed SGD weights, $\theta_i'$ over epochs and show that the errors are amplified, suggesting instability. We use the average relative L1 absolute difference between weights: $\mbox{RelL1}(\theta,\theta'):=\frac{2}{N}\sum_i |\theta_i-\theta_i'| / ( |\theta_i|+|\theta_i'|)$, where $N$ is the number of network parameters. This measure is used as the numerical error is multiplicative and relative to the weight. Note that $\mbox{RelL1}$ is bounded by 2. Results are shown for multiple networks in Figure~\ref{fig:iteration} (left). The difference rapidly grows, indicating errors are amplified, and then the growth is restrained at larger epochs. Although the errors are restrained, the initial error growth is enough to result in different test accuracies. This provides evidence of a restrained rather than global instability.

{\bf Evidence Divergence is Due to Instability}: We provide evidence that this divergence phenomenon is due to instability by showing that decreasing the learning rate to a small enough value can eliminate the divergence, even initially. We experiment with various fixed learning rates. We use Resnet56 under the same settings as before. For smaller learning rates, more iterations would have to be run to move an equivalent amount for a larger rate. We thus choose the iterations for each run such that the learning rate times the number of iterations is constant. The result is shown in Figure~\ref{fig:iteration} (right), where the axis is scaled as mentioned. When the learning rate is chosen small enough, the floating point errors are attenuated over iterations, indicating the process is stable (note it is not stuck at a local minimum - see Appendix~\ref{app:extra_tables}). Above this level, the errors are amplified.

\section{PDE Stability Analysis of Simplified CNNs}

In this section, we use numerical PDE tools introduced in Section~\ref{sec:background} to analyze a one-layer (and then multi-layer) CNN to show analytically how the instability phenomenon described in the previous section can arise. To do this, we start with a continuum model of the network, derive the analytical formula for the gradient descent, which is a PDE, discretize it, which serves as a model for the optimization algorithm, and analyze the stability of the discretization through Von-Neumann analysis. 
This approach where the continuum model is considered, although different than the typical approach of analyzing the final optimization scheme directly, allows us to bring insights from PDEs to reason about the discretized optimization, which provides a complimentary understanding to analyzing the final optimization directly.

\subsection{Network, Loss and Gradient Flow PDE}
Let $I : \R^2 \to \R$ be an input image to the network, $K : \R^2 \to \R$ be a convolution kernel (we assume $I$ and $K$ to be of compact support and bounded), $r: \R\to\R$ be the Swish activation, and $s : \R\to\R$ is the sigmoid function. We consider the following CNN, is a simplified version of VGG:
\begin{equation}
f(I) = s\left[ \int_{\R^2} r(K \ast I)(x) \ud x \right].
\end{equation}
This network consists of a convolution layer, an activation layer, a global pooling layer, and a sigmoid layer. As in deep learning packages, $\ast$ will denote the cross-correlation (which we call convolution). The bias is not included in the network as it does not impact our analysis nor conclusions.

We assume a dataset that consists of just a single image $I$ and its label $y\in \{0,1\}$. This is enough to demonstrate the instability phenomena. We consider the following regularized cross-entropy loss:
\begin{equation} \label{eq:loss}
  L(K) = \ell( y, \hat y ) + \frac 1 2 \alpha \|K\|^2_{\mathbb{L}^2}, \quad  \mbox{where} \quad \hat y = f(I),
\end{equation}
$\ell$ denotes cross-entropy, the second term is regularization ($\mathbb L^2$ norm squared of the kernel), and $\alpha>0$ is a parameter (weight decay). Our conclusions are not restricted to cross-entropy loss (as will be seen in the Appendix~\ref{app:derivations}), but it makes some expressions simpler. We now compute the continuum limit of the gradient descent of the loss \eqref{eq:loss}, which results in a PDE called the gradient descent PDE. We now let $K$ be parameterized by time $t$, which is the continuum limit of the discrete iteration, to describe the path of optimization (evolution), so that $K : [0,\infty)\times \R^2\to\R$. The gradient descent PDE of \eqref{eq:loss} is given by (see the Appendix~\ref{sec:derivation_nonlinear} for proof):
\begin{thrm}[Gradient Descent PDE of the Loss \eqref{eq:loss}]
    The gradient descent PDE with respect to the loss \eqref{eq:loss} is 
    \begin{equation}
    \label{eq:PDE_no_constraint}
     \partial_t K = -\nabla_K L(K) = -(\hat y-y)r'(K \ast I) \ast I - \alpha K,
    \end{equation}
    where $r'$ denotes the derivative of the activation, and $\partial_t$ denotes the partial derivative with respect to time.
    If we impose the constraint that $K$ has compact support ($K$ is zero outside $[-w/2,w/2]^2$), then the constrained (or projected) $\mathbb{L}^2$ gradient descent is given by 
    \begin{equation} \label{eq:PDE_constraint}
      \partial_t K = \left[ -(\hat y-y)r'(K \ast I) \ast I - \alpha K\right] \cdot W,
    \end{equation}
    where $W$ is a windowing function ($1$ inside $[-w/2,w/2]^2$ and zero outside).
\end{thrm}
Note that if we consider $K$ to be a sum of variably weighted indicator functions with compact support centered around the pixels (an approximation of the discrete case), then the gradient descent w.r.t the weights is the evaluation of the right hand side of \eqref{eq:PDE_no_constraint} at the location of the indicator functions. Studying the continuum limit (as the resolution increases by increasing the number of indicator functions) gives insight into the discrete case.

\subsection{Stability Analysis of Optimization}
\label{sec:stability}
We now analyze the stability of the standard discretization of the gradient descent PDEs. Note that the PDEs are non-linear, and in order to perform the stability analysis, we study linearizations of the PDE around the switching regime of the activation, i.e., $K\ast I=0$.  As shown in the next sub-section, away from the transitioning regime, a similar analysis applies.

We first study the linearization of \eqref{eq:PDE_no_constraint}, and then we adapt the analysis to \eqref{eq:PDE_constraint}. 
\cut{
We recall the definition of the Swish function:
\begin{equation}
  r(x) = \frac{x}{1+e^{-\beta x}},
\end{equation}
where $\beta>0$ is a hyper-parameter such that when $\beta \to \infty$, $r$ approaches the rectified linear unit. 
}
The linearization of the PDE results in the following (see Appendix~\ref{sec:derivative_linear} for the derivation):
\begin{thrm}[Linearized PDE]
    The linearization of the PDE \eqref{eq:PDE_constraint} around $K\ast I=0$ is 
    \begin{equation} \label{eq:linear_PDE}
        \partial_t K = \left[ 
        -\frac a 2 (\bar I + \beta [(K\ast I) \ast I] ) - \alpha K
        \right] W, \quad \mbox{ where }\quad  a = s' \partial_{\hat y }\ell = \hat y - y,
    \end{equation}
    $a$ is considered constant with respect to $K$, $\bar I$ denotes the integral (sum) of the values of $I$ (assumed finite support), and $\beta>0$ is the parameter of the Swish activation, $r(x) := \frac{x}{1+e^{-\beta x}}$.
    \cut{
    We consider two models for $a$:
    \begin{itemize}
    \item $a$ is a constant
    \item $a$ is linearized: $a = s(0) - y + 0.5 s'(0) \bar K \bar I$
    \end{itemize}
    }
\end{thrm}
In Appendix~\ref{app:derivations}, we also consider a non-constant linear model of $K$ for $a$. This leads to similar conclusions so we present the constant model for simplicity.

\cut{We now analyze the stability of the linearized equation using Von-Neumann analysis.}
We discretize the linearization, which results in a discrete optimization algorithm, equivalent to the algorithm employed in standard deep learning packages. Using forward Euler discretization gives
\begin{equation}\label{eq:discrete}
  K^{n+1} - K^n = \left[ -\frac a 2 \Delta t ( \bar I + \beta [(K^n\ast I) \ast I] ) - \Delta t \alpha K^n\right] W,
\end{equation}
where $n$ is the iteration number, and $\Delta t$ denotes the step size (learning rate). To derive the stability conditions, we compute the spatial DFT of \eqref{eq:discrete} (see Appendix~\ref{app:derivations} for details):
\begin{thrm} [DFT of Discretization]
    The DFT of the discretization \eqref{eq:discrete} of the linearized PDE is
    \begin{equation} \label{eq:DFT}
    \hat K^{n+1} - \hat K^n  = -\frac a 2 \Delta t ( \bar I + \beta \hat K^n |\hat I|^2 ) \ast \left[\prod_{j=1}^2 \sinc{\left(\frac w 2 \omega_j\right)}
    \right]
    - \Delta t \alpha \hat K^n,
    \end{equation}
    where $\hat K^n$ denotes the DFT of $K^n$, and $\sinc$ denotes the sinc function. In the case that $w\to \infty$ (the window support becomes large),
    the DFT of $K$ can be written in terms of the amplifier $A$ as
    \begin{equation} \label{eq:DFT_amplifier}
      \hat K^{n+1}(\omega)  = A(\omega) \hat K^n(\omega)
                                       -\frac a 2 \Delta t  \bar I,\,\, \mbox{ where }\,\,
      A(\omega) = 1 -\Delta t \left(\alpha + \frac 1 2 a\beta |\hat I(\omega)|^2\right).
    \end{equation}
\end{thrm}

In order for the updates to be a stable process, the magnitude of the amplifier should be less than one, i.e., $|A| < 1$. This results in the following stability criteria  (see the Appendix~\ref{app:derivations} for a derivation):
\begin{thrm}[Stability Conditions]
    The discretization of the linearized PDE \eqref{eq:PDE_no_constraint} whose DFT is given by \eqref{eq:DFT} is stable (in the case $w\to \infty$, i.e., the window gets large) if and only if the following conditions on the weight decay $\alpha$ and the step size (learning rate) $\Delta t$ are met:
    \begin{align} \label{eq:alpha_bounds}
        \alpha < \alpha_{max} := \frac{2}{\Delta t} - \frac 1 2 a \beta \min_{\omega} |\hat I(\omega)|^2\,\,
        \mbox{ and }\,\,
        \alpha > \alpha_{min} := - \frac 1 2 a \beta \max_{\omega} |\hat I(\omega)|^2.
    \end{align}
\end{thrm}

Note the sign of $a = \hat y - y$ can be either positive or negative; in the case $a>0$, the second condition is automatically met (since $\beta>0$). However in the case that $a<0$, the weight decay must be chosen large enough to be stable. The first condition shows that the weight decay and the step size satisfy a relationship if the optimization is to be stable. Note that the conditions also depend on the structure of the network, the current state of the network (i.e., through the dependence of $a$), and the frequency content of the input. This is different than current deep learning practice, where the weight decay is typically chosen constant through the evolution and does not adapt with the state of the network, and the step size is typically chosen empirically without dependence on the weight decay.

We now extend the previous stability analysis from the gradient descent to Nesterov accelerated gradient descent, which is widely used in deep neural network training. As we shall see, we arrive at similar conclusions (CFL conditions that result in upper and lower bounds on $\alpha$) as the gradient PDE. We first note the underlying PDE to Nesterov's method, which enables the use of Von Neuman Analysis and then state the CFL conditions.

\cut{
We start with the continuum PDE, discretize it - resulting in Nesterov's scheme, and then analyze the stability. The continuum equivalent PDE for the optimization of a loss function with momentum is given by (see \citep{benyamin2020accelerated}):
\begin{equation}\label{eq:pde_momentum}
    \partial_{tt} K + d \cdot \partial_t K = -\nabla L(K),
\end{equation}
where $\partial_{tt}$ denotes the second derivative in time, $d>0$ is a constant that denotes the damping coefficient,  and $u$ denotes the evolving variable of optimization. We may use a semi-implicit Euler style discretization of the above PDE that results in classic two-part Nesterov recursion (see \citep{benyamin2020accelerated}).
}

\begin{thrm}[Nesterov Accelerated Gradient Descent and its Underlying PDE]
The continuum limit of Nesterov accelerated gradient descent for a loss $L$ is
\begin{equation}\label{eq:pde_momentum}
    \partial_{tt} K + d \cdot \partial_t K = -\nabla L(K),
\end{equation}
where $\partial_{tt}$ denotes the second derivative in time, $d>0$ is a constant that denotes the damping coefficient, and $u$ denotes the evolving variable of optimization. The semi-implicit Euler discretization for the PDE \eqref{eq:pde_momentum} is a two-point recursion resulting in Nesterov's scheme:
\begin{align} 
\label{eq:semi_implicit_scheme}
V^n &= K^n + \frac{2 - d\Delta t}{2 + d\Delta t}(K^n - K^{n-1})
 \\
 \label{eq:semi_implicit_scheme2}
 K^{n+1} &= V^n - \frac{2\Delta t^2}{2 + d\Delta t}\nabla_V L(V^n)
\end{align}
where  $V^n$ is the partial update.
\end{thrm}

For a proof, we refer the reader to \citep{benyamin2020accelerated} and more discussion in Appendix~\ref{sec:nesterov} (and \citep{su2014differential,wibisono2016variational} for variational approaches to accelerated gradient methods). We use Von Neumann analysis similar to the analysis of gradient descent to arrive at the following stability conditions (detailed proof is in the Appendix~\ref{sec:nesterov}):
\begin{thrm}[Stability Conditions for Nesterov Gradient Descent]
    \label{thrm:nesterov_bound}
    The semi-implicit scheme of the linearized momentum PDE \eqref{eq:semi_implicit_scheme}, \eqref{eq:semi_implicit_scheme2} for the loss \eqref{eq:loss} is stable (in the case $w\to \infty$) if and only if the following conditions on the weight decay $\alpha$ and the step size $\Delta t$ are met:
    \begin{align} \label{eq:nesterov_alpha_bounds}
        \alpha &< \alpha_{max} := \frac{4}{3\Delta t^2} - \frac 1 2 a \beta \min_{\omega} |\hat I(\omega)|^2 
        \,\,\mbox{ and }\,\,
        \alpha > \alpha_{min} := - \frac 1 2 a \beta \max_{\omega} |\hat I(\omega)|^2.
    \end{align}
\end{thrm}
\noindent Thus, we see that Nesterov momentum does not change the existence of bounds relating step size and regularization, though Nesterov's scheme has more generous step sizes.

\subsection{Empirical Validation of Stability Bounds for Linear PDE}
\label{sec:empirical_bound_PDE}
\begin{figure}
\floatsetup{heightadjust=all, valign=c}
\begin{floatrow}
\ffigbox[0.55\textwidth][]{%
    \includegraphics[width=0.8\columnwidth]{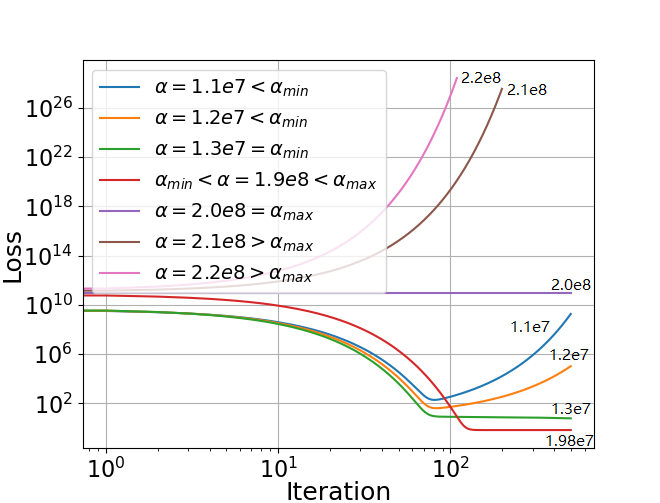}
}{%
  \captionof{figure}{Empirical validation of stability bounds for the linearized PDE \eqref{eq:linear_PDE}. When the weight decay is chosen such that $\alpha \in (\alpha_{min}, \alpha_{max})$, the PDE is stable and the loss converges, otherwise, the PDE can be unstable and diverge. The pink and brown lines are clipped as the loss exceeded the largest float.}%
  \label{fig:linear_PDE_stability_plot}%
}%
\capbtabbox[0.43\textwidth]{%
    \small
  	\begin{tabular}{ccccc}  
	\toprule
    Kernel & $\alpha_{min}^e$ & $\alpha_{min}$ & $\alpha_{max}^e$ &  $\alpha_{max}$\\
    \midrule
    16x16 & 3.8e6 & 1.07e9 & 2e8 & 2e8\\
	\midrule
	32x32 & 1.3e7 & 1.07e9 & 2e8 & 2e8\\
	\midrule
    64x64 & 3.8e7 & 1.07e9 & 2e8 & 2e8\\
	\bottomrule
	\end{tabular}
}{%
  \captionof{table}{Comparison of the bounds on weight decay, $\alpha$, between the non-windowed linear PDE, i.e., $\alpha_{min}$ and $\alpha_{max}$ in \eqref{eq:alpha_bounds} and the windowed linear PDE in \eqref{eq:linear_PDE} found empirically, i.e., $\alpha_{min}^e$ and $\alpha_{max}^e$. Verifies bounds exist in windowed case.}%
  \label{tab:alpha_window}
}
\end{floatrow}
\end{figure}

Figure~\ref{fig:linear_PDE_stability_plot} empirically validates the existence of upper and lower bounds on $\alpha$. We chose the learning rate $\Delta t = \mbox{1e-8}$, $a=-0.5, \beta=1$, initialize the weights $K$ ($32\times 32$) according to a normal distribution (mean 0, variance 1), and $I$ is a $256 \times 256$ checkerboard pattern filled with 1 and -1.

The above stability conditions were under the condition that the windowing function, $W$, had infinite support. Thus, the evolution could cause the initial kernel's (finite) support to grow over iterations, which is not representative of fixed, finite support kernels used in deep network practice. In the case that $W$ is finite and fixed support, it is not possible to analytically find a simple multiplicative amplifier factor as in \eqref{eq:DFT_amplifier} since the convolution with the sinc function in \eqref{eq:DFT} does not separate from $\hat K^n$. In the case that $\hat K^n$ is a constant (i.e., $K^n$ is a discrete delta function or support of size 1, a single point), $(\hat K^n |\hat I|^2) \ast$ $\prod_{j=1}^2 \sinc{\left(\frac w 2 \omega_j\right)}$ $= \hat K^n (|\hat I|^2 \ast \prod_{j=1}^2 \sinc{\left(\frac w 2 \omega_j\right)}$), in which case one can write the amplifier function as
\begin{equation}
     A(\omega) = 1 -\Delta t \left(\alpha + \frac 1 2 a\beta |\hat I(\omega)|^2 \ast \prod_{j=1}^2 \sinc{\left(\frac w 2 \omega_j\right)} \right),
\end{equation}
which has the effect that the upper bound for $\alpha$ increases and the lower bound decreases in \eqref{eq:alpha_bounds}, but maintains the existence of upper and lower bounds. 

We conjecture this is also true when $K^n$ has support larger than $1$, and the bounds on $\alpha$ converge to \eqref{eq:alpha_bounds} as the support of the kernel is increased. We provide empirical evidence. We find the lower and upper bounds for $\alpha$ by running the PDE and noting the values of $\alpha$ when the scheme first becomes unstable. We compare it to the bounds in \eqref{eq:alpha_bounds}. We use the same settings as the previous experiment. See results in Table~\ref{tab:alpha_window}, which verifies the bounds approach of the infinite support case.

\subsection{Restrained Instabilities in the Non-linear PDE}
\label{subsec:restrained_instabilities}

\cut{
In the previous sections, we analyzed the gradient descent PDE around $K=0$ in which the linearization is valid.  We showed that in this regime, the standard discretization for the gradient descent PDE is stable only under conditions on the weight decay and learning rate. 
}

We now explain how the linear analysis around $K\ast I=0$ in the previous sections applies to analyzing full non-linear PDE. In particular, we show that the non-linear PDE is also not stable unless similar conditions on the weight decay and step size are met. However, the non-linear PDE can transition to regimes that have more generous stability conditions where it could be stable, and the PDE can move back and forth between stable and unstable regimes, resulting in restrained instabilities.

{\bf Approximation of Non-Linear PDE by Linear PDEs and Stability Analysis}: The non-linear PDE could move between three different regimes: when the activation is ``activated'' ($r'(x)=1$), ``not activated'' ($r'(x)=0$), and ``transitioning'' ($r'(x)$ is a non-constant linear function). In the transitioning regime, $K\ast I \approx 0$, in which case the linear analysis of the previous section applies, and in this regime, there are conditions on the weight decay and step size for stability. In the not-activated regime, $K\ast I$ is negative and away from zero, the non-linear PDE is driven only by the regularization term, which is stable when $\alpha <  \frac{2}{\Delta t}$. This is typically true in current deep learning practice where weight decay is chosen on the order of 1e-4 and the learning rate is on the order of $<$1e-1. In the activated regime, $K\ast I$ is positive and away from zero and the non-linear PDE can be approximated by a linear PDE (see Appendix~\ref{subsec:linear_activate}), which gives the stability condition: $\alpha < \frac{2}{\Delta t}-\frac{\bar I^2}{8}$, similar to the stability condition in the transitioning regime.

{\bf Formation of Restrained Instabilities}: Note that different regions in the kernel domain could each be in different regimes and therefore governed by different linear PDEs discussed earlier. See Table~\ref{tab:regime_plot} \footnote{ Note that the input to $r$ is a function $K\ast I$ that can be positive, negative or zero at each point in its domain so one should treat $K\ast I$ point-wise. However, doing so because we are considering linearizations, results in analyzing the same three PDEs in Table~\ref{tab:regime_plot}. To see this, note that $
r'(K\ast I)\ast I \approx 
\mathbf{1}_{K\ast I > \varepsilon}\ast I + 
[ (K\ast I)\mathbf{1}_{|K\ast I| \leq \varepsilon} ] \ast I$, where $\mathbf{1}$ denotes the characteristic function. If we linearize around a point, then the linear approximation of the characteristic function is just the constant function with the constant being the characteristic function at that point. This yields the linear PDEs considered in Table~\ref{tab:regime_plot} for points $x$ such that $K\ast I(x)>0$, $K\ast I(x) \approx 0$, $K\ast I(x) <0$, respectively.
}. 
Such regions could then switch between different regimes throughout the evolution of the PDE, and thus the evolution could go between stable and unstable regimes. For example, in the transitioning regime, the instability quickly drives the magnitude of the kernel up (if the stability conditions are not met) and thus the region to a possibly stable (e.g., activated) regime. Data-driven terms could then kick the kernel back into the (unstable) transitioning regime, and this switching between regimes could happen indefinitely. As a result, instabilities occur but can be ''restrained'' by the activation.  The optimization nevertheless amplifies and accumulates small noise in the transitioning regime.

\cut{
\begin{figure}
    \centering
    \floatbox[{\capbeside\thisfloatsetup{capbesideposition={left,top},capbesidewidth=0.4\textwidth}}]{figure}[\FBwidth]{\caption{The non-linear PDE is approximated by three linear PDEs in different regions of the tensor (kernel) space ("activated", "not activated" and "transitioning"). The non-linear PDE can transition between these regions in which different linear PDEs approximate the behavior.} \label{fig:regime_plot}
    }{\includegraphics[width=0.5\textwidth]{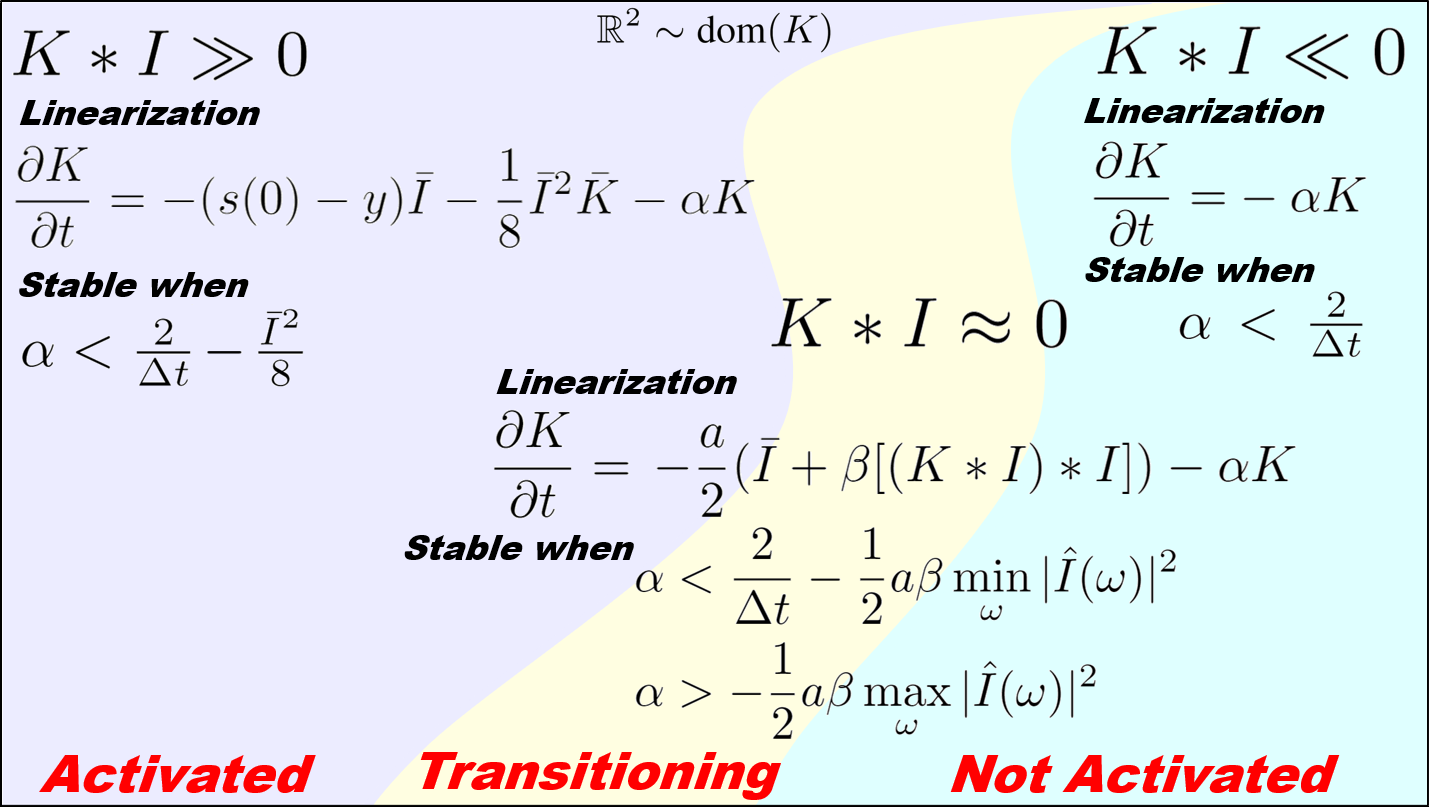}}
\end{figure}
}

\begin{table}
    \centering
    \small
    {\renewcommand{\arraystretch}{1.3}
    \begin{tabular}{c|c|c|c}
         &  Activated & Transitioning & Not Activated \\\hline
        Region & $\{ x: K\ast I(x) \gg 0\}$ & $\{ x : K \ast I(x) \approx 0\}$ &
        $\{ x : K\ast I(x) \ll 0 \}$ \\\hline
        \multirow{2}{*}{Linear PDE}
        & $\partial_t K = -(s(0)-y)\bar I $ 
        & $\partial_t K = -\frac{a}{2}(\bar I + \beta[(K\ast I)\ast I])$ 
        & \multirow{2}{*}{$\partial_t K = -\alpha K$} \\
        & $- \frac 1 8 \bar I^2\bar K -\alpha K$ 
        & $-\alpha K$ & \\\hline
        \multirow{2}{*}{Stability} & 
        \multirow{2}{*}{$\alpha < \frac{2}{\Delta t}-\frac{\bar I^2}{8}$} &
        $\alpha < \frac{2}{\Delta t} - \frac{1}{2} a \beta \min_{\omega} |\hat I (\omega)|^2 $ &  
        \multirow{2}{*}{$\alpha < \frac{2}{\Delta t}$} \\
         & & $\alpha > -\frac{1}{2} a\beta \max_{\omega} |\hat I(\omega)|^2$ & \\
    \end{tabular}%
    }
    \caption{The non-linear PDE is approximated by three linear PDEs in different regions of the tensor space ("activated", "not activated" and "transitioning"). The non-linear PDE can transition between these regions in which different linear PDEs approximate the behavior.}
    \label{tab:regime_plot}
    
\end{table}

\cut{
We choose the image $I$ to be of size $256\times256$ with checkerboard pattern (so the image contains a single frequency and $\bar I=0$) and its label to be $y=1$. We choose the kernel $K$ to be $32\times32$, which we initialize with a random Gaussian initialization (mean 0, variance 1). 
}

\begin{figure}[h]
\begin{minipage}[h]{0.5\linewidth}
\centering
\includegraphics[width=0.8\columnwidth,clip,trim=0 0 10 25]{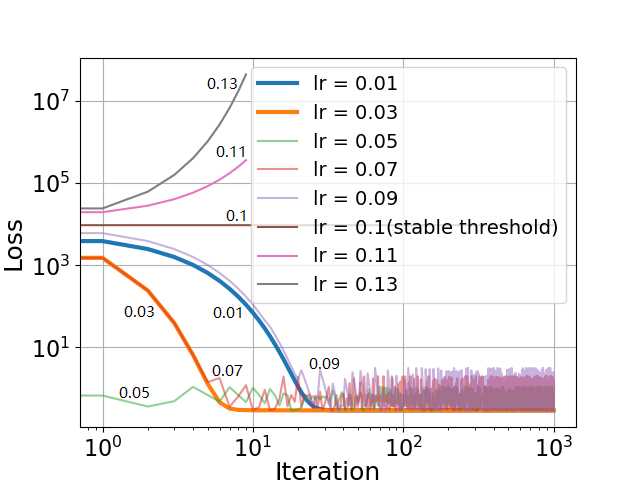}%
\end{minipage}%
\begin{minipage}[h]{0.5\linewidth}
\centering
\includegraphics[width=0.8\columnwidth]{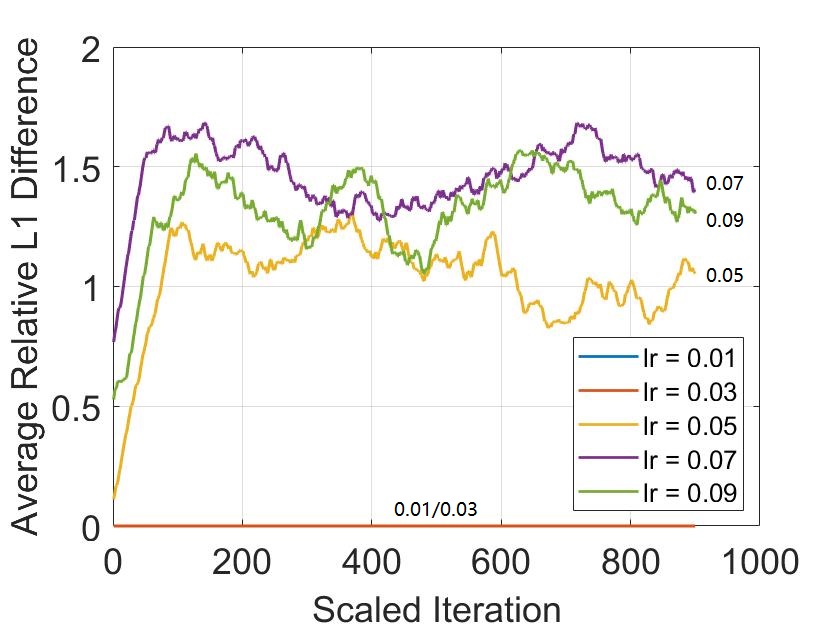}%
\end{minipage}

\caption{{\bf [Left]}: Loss vs iterations of the non-linear PDE \eqref{eq:PDE_constraint} for various choices of learning rates ($\Delta t$). The loss for various learning rates are consistent with the theory predicted for fully stable, fully unstable and restrained instabilities. \cut{When $\Delta t > 0.1$ all regimes are unstable. When $\Delta t < 0.03$, all regimes are stable. When $0.03\leq \Delta t < 0.1$, the PDE can transition indefinitely between the ``transitioning" regime (unstable) and the ``activated" regime (stable).  This results in oscillations (``restrained" instabilities) as verified empirically in this plot.} {\bf [Right]}: L1 error accumulation in the non-linear PDE. The plot is consistent with expected error accumulation for restrained instabilities, and fully stable regimes. }
\label{fig:oscillation_plot}
\end{figure}

{\bf Empirical Validation}: We now empirically demonstrate that restrained instabilities are present and error amplification occurs in the (non-linear) gradient descent PDE of the one-layer CNN even when the dataset consists of a single image.  We choose $\alpha=20, \beta=1, y=1$ and the same input image $I$ and kernel $K$ initialization as specified in Section~\ref{sec:empirical_bound_PDE}. Figure~\ref{fig:oscillation_plot} (left) shows the loss function over iterations for various learning rate ($\Delta t$) choices, and the right plot shows the evolution of the L1 relative difference between weights of SGD ($k=1$) and perturbed SGD ($k=3$) as discussed in Section~\ref{sec:floatingpoint_error}. From the previous discussion and \eqref{eq:alpha_bounds}, we know that it is necessary that $\Delta t < 0.1$ in order for all regimes to be stable. This is confirmed by the plot on the left: when $\Delta t > 0.1$, the loss blows up. When $\Delta t < 0.03$, which is the empirically determined threshold for the transitioning region, all regimes are stable, and the kernel converges as shown in the plot. When $0.03 < \Delta t \leq 0.1$, the PDE transitions between the (unstable) transitioning regime and the stable one (activated), which introduces oscillations.  This is consistent with our theory outlined previously. On the plot on the right of Figure~\ref{fig:oscillation_plot}, we show divergence caused by the floating perturbations introduced in Section~\ref{sec:floatingpoint_error} for various $\Delta t$. When  $0.03 < \Delta t \leq 0.1$ (in the restrained instability regime), the error builds up quickly and then levels off. When $\Delta t < 0.03$, errors are not amplified. This is consistent with the behavior of multi-layer networks in Section~\ref{sec:floatingpoint_error} validating our theory.

\cut{\textcolor{red}{{\bf Kernel Demonstration}: We now illustrate the restrained instability in the kernel space. We use the same experimental setting as the last part, but we switch the input image to the natural image used in Figure \ref{fig:beltrami-1}. In Figure \ref{fig:kernel}, we see examples of pure stability, restrained instability, and global instability. For restrained instability, we could see oscillation on a global scale. The kernel is switching among several states.}
}

\subsection{Restrained Instabilities in Multi-layer Networks}
\label{subsec:multilayer_restrained}
The previous sub-sections formulated a theory on how the divergence phenomena for practical deep network optimization reported in Section~\ref{sec:floatingpoint_error} might arise. We found that even the simplest (one-layer) CNN exhibits these phenomena through analytical techniques. In multi-layer networks, there are more possibilities for such restrained instabilities to arise in multiple layers and multiple kernels of the network. Such instabilities may arise only in some kernels/layers. The analysis in the previous sub-section, though only for a single layer, is also relevant to multi-layer networks. In particular, 
we may isolate each layer in a multi-layer network and treat
the output of the previous layer as input to the given layer, and analyze the stability of that one-layer sub-network in isolation (which the analysis in the previous sub-sections applies). We next give analytic justification to this intuition.

We consider a model for an $N$-layer binary classification convolutional neural network by stacking $N$ convolutional and activation layers, followed by a global pooling and sigmoid layers as follows:
\begin{align}
    f(I) &= s\left[ \int F(I)(x)\ud x  \right], \\
    \quad F(I) &= F_N\circ F_{N-1} \circ \cdots \circ F_1(I),\,\,
    \mbox{where}\,\,
    F_i(I_i) = r(K_i\ast I_i),
\end{align}
$K_i$ is the convolutional kernel for the $i^{\text{th}}$ layer, and $I_i = F_i\circ \cdots \circ F_1(I)$ is the input to the $i^{\text{th}}$ layer. We consider the following generalization of the loss function considered in the one-layer case:
\begin{equation} \label{eq:loss_multilayer}
L(K_1,\ldots, K_N) = \ell(y,\hat y) + \frac 1 2 \alpha \sum_j \| K_j \|_{\mathbb{L}^2}^2, \quad \hat y = f(I).
\end{equation}
We now derive the expression for the gradient of the loss with respect to the kernels (see Appendix~\ref{app:multilayer}):
\begin{thrm}[Multi-layer CNN Kernel Gradient] \label{thrm:mulilayer_gradient}
    The gradient of the loss \eqref{eq:loss_multilayer} with respect to the kernel $K_i$ is given by 
    \begin{equation} \label{eq:grad_multi_layer}
        \nabla_{K_i} L = 
        (\hat y - y) \left[ \overline{\nabla_{I_{i+1}} F_{N,i+1}(I_{i+1})} \cdot r'(K_i\ast I_i) \right] \ast I_i + \alpha K_i,
    \end{equation}
    where
    \begin{equation}
        \overline{\nabla_{I_{i+1}} F_{N,i+1}(I_{i+1})}(x) = \int \nabla_{I_{i+1}} F_{N,i+1}(I_{i+1})(z,x) \ud z,
    \end{equation}
    $F_{N,i+1}= F_N\circ F_{N-1} \circ \cdots \circ F_{i+1}$, and $\nabla_{I_{i+1}} F_{N,i+1}$ is the gradient with respect to the input to the $(i+1)^{\text{th}}$ layer, $I_{i+1}$.
\end{thrm}

Note the similarity of the gradient expression \eqref{eq:grad_multi_layer} to the gradient of the one-layer case \eqref{eq:PDE_no_constraint}. This shows that stability analysis of the one-layer network is representative of the multi-layer network since we will show that $\nabla_{I_{i+1}} F_{N,i+1}(I_{i+1})$ is approximately independent of $K_i$. Therefore, instability in the multi-layer case must arise at some layer for which the one-layer analysis applies. Optimization will be stable when stability conditions are met for each layer.

Now, we justify that $\nabla_{I_{i+1}} F_{N,i+1}(I_{i+1})$ is independent of $K_i$, and hence the linearization of the gradient \eqref{eq:grad_multi_layer} is thus the same as the one-layer case. As in the one-layer case, we consider three different regimes based on the state of the activation: $K_i\ast I_i \approx 0$, $K_i\ast I_i \ll  0$, and $K_i\ast I_i \gg 0$. In these cases, one can show (see Appendix~\ref{app:multilayer}):
\begin{equation}
     \nabla_{I_{i+1}} F_{i+1}(x,z) =  
     K_{i+1}(z-x)
        \begin{cases}
            \frac 1 2  & K_i\ast I_i \approx 0, K_i\ast I_i \ll 0 \\
            1 & K_i\ast I_i \gg 0
        \end{cases}.
\end{equation}
Therefore, the gradient at the $i+1$ layer approximately ceases dependence on $K_i$. We may apply a similar analysis to argue that $\nabla_{I_{i+k}} F_{i+k}(I_{i+k})$ for each $k$ is independent of $K_i$. Thus, the gradient $\nabla_{I_{i+1}} F_{N,i+1}(I_{i+1})$ is also independent of $K_i$. Thus, the linearization follows the one-layer case. The stability analysis can be approximated by approximating $\nabla_{I_{i+k}} F_{N,i+1}(I_{i+1})$ with a constant. Stability could be ensured by choosing the constant with the most stringent stability conditions. With this approximation, the stability analysis is exactly the same as in the previous sub-sections, however, the constant $a$ is replaced with $\hat a \gamma$, where $\gamma$ is the constant approximation of $\overline{\nabla_{I_{i+k}} F_{N,i+1}(I_{i+1})}$, which changes the values of the stability bounds, but not the nature. 

The analysis shows restrained instabilities in multi-layer networks arise in a similar fashion to how they may arise in a single-layer network. The analysis shows that restrained instabilities arise locally at a layer at a given location and could spread to other layers through the network structure.

\cut{
We have illustrated stable, unstable, and restrained instabilities (e.g., oscillations) in the case of a one-layer network. Similar phenomena may exist in standard training procedures of deep networks, where there may be more possibilities for the optimization to be in all these regimes as multi-layer networks have many kernels across many layers. Indeed in the experiment on Resnet56 in Figure~\ref{fig:iteration}, it may be the case that several kernels are in the unstable regime initially when the learning rate is high, and the numerical errors build up, and then as the learning rate is lowered, it could go into contained instabilities or stable regions.
}

\section{Restrained Instabilities as an Explanation for the Edge of Stability}

In recent work \citep{DBLP:journals/corr/abs-2103-00065}, a phenomenon called the ``Edge of Stability'' (EoS) was reported. It was empirically observed that deep learning optimization occurs in a regime that is slightly beyond the stability condition predicted by classical optimization theory in which a quadratic approximation of the loss function is used. Classical theory predicts that $\mbox{lr} < 2/\lambda_{max}$ for stability, where $\lambda_{max}$ is the maximum value of the Hessian of the loss function, called \emph{sharpness}. For learning rates chosen in practice, the sharpness slightly hovers above the predicted $2/\mbox{lr}$ bound. Nevertheless, optimization seemingly does not diverge and gives practically meaningful results, suggesting possibly a new optimization theory is needed. In this section, we empirically show that restrained instabilities reported in this paper may offer an explanation to the EoS, and further that our theory implies additional predictions to the EoS not previously known. 

At a conceptual level, it would not be unexpected that restrained instabilities, which result in oscillations, are at the edge of stability as between stable and fully unstable regimes, oscillations can arise. Consider for example, $f(x)=x^2$ whose gradient descent oscillates between 1 and -1 when the step size is chosen to be 1, which is at EoS.  In the case of neural networks, restrained instabilities represent a continuing oscillation of a nonlinear dynamical system between regimes where its linearization is classically stable (where errors are attenuated as the system evolves) and regimes where its linearization is classically unstable (where errors grow exponentially as the system evolves). This phenomenon can arise when the deregularization effect of instabilities that begin to grow in an unstable regime cause the nonlinear system to transition into a regime where a regularization effect takes over, which begins to reverse the deregularized features and re-transition the system back into the original unstable regime. This process repeats itself and thereby causes an oscillatory behavior in which the system therefore "hovers" right around the transition point of instability. In the case of EoS, \citep{DBLP:journals/corr/abs-2103-00065} phenomenologically observed that for large enough learning rates, the largest eigenvalue of the network Hessian would hover around $2/\mbox{lr}$. They connect this in turn, with hovering around a threshold of stability (hence the term Edge of Stability) in the case of a simple quadratic model. While this doesn't explain why the instability occurs in the first place, nor why it hovers there rather than continues to diverge with an ever-growing eigenvalue, it does nevertheless demonstrate that the dynamical system is transitioning back and forth across this detectable indicator of stability. In this manuscript, by considering the network in its continuum limit (PDE), we have now offered an explanation, through Von Neumann stability analysis, which traces precisely where this stable/unstable transition point arises in discretized gradient descent, so that the EoS is now both predictable and explainable, rather than just observable, in terms of the chosen learning rate.

\subsection{Restrained Instabilities are at the Edge of Stability}

\begin{figure}[h]
\begin{minipage}[h]{0.5\linewidth}
\centering
\includegraphics[width=0.8\columnwidth]{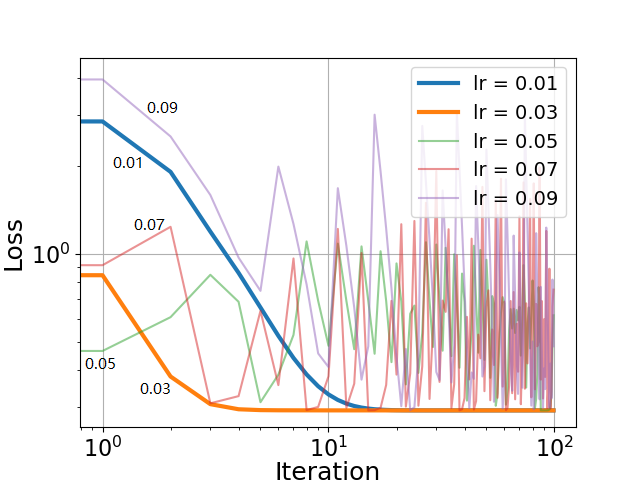}%
\end{minipage}%
\begin{minipage}[h]{0.5\linewidth}
\centering
\includegraphics[width=0.8\columnwidth]{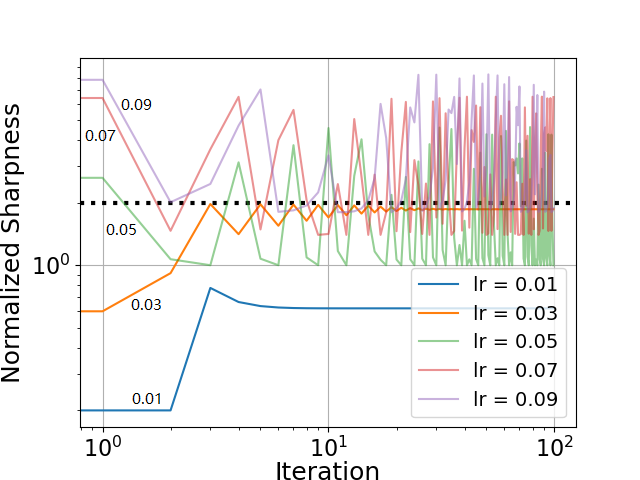}%
\end{minipage}

\caption{{\bf [Left]}: Loss vs iterations plot for the discretization of the non-linear gradient descent PDE \eqref{eq:PDE_constraint}. {\bf [Right]}: Normalized Sharpness (sharpness * lr) vs iterations for same learning rates in the left plot. For stable learning rates (0.01, 0.03), the normalized sharpness is always below 2 (marked by the horizontal dashed line). However, for learning rates (0.05, 0.07, 0.09) for which restrained instabilities (oscillations) occur, normalized sharpness oscillates and can exceed the classical stability bound of 2.}
\label{fig:sharpness_plot}
\end{figure}

We empirically demonstrate that restrained instabilities occur at the EoS, providing empirical validation that our theory can explain EoS. To show this, we consider the simple 1-layer CNN in Section~\ref{subsec:restrained_instabilities} and repeat the experiment for the non-linear PDE \eqref{eq:PDE_constraint} under exactly the same settings, in particular with one training datum. We numerically compute and plot the normalized sharpness, $\mbox{lr} \cdot \lambda_{max}$, across iterations for various learning rates in Figure~\ref{fig:sharpness_plot}. We observe that the learning rates predicted from our theory when the restrained instabilities occur are also at the EoS, i.e., when the normalized sharpness exceeds 2. Learning rates predicted from our theory when the non-linear PDE is fully stable is also when the normalized sharpness is less than the EoS boundary of normalized sharpness being 2. This suggests that restrained instabilities and our numerical PDE theory offer an explanation for the EoS.

\subsection{New Insights on the Edge of Stability Predicted by Our Theory}

We discuss new insights relating to the EoS not known before that our theory reveals.

\begin{figure}[h]
\begin{minipage}[h]{0.5\linewidth}
\centering
\includegraphics[width=1\columnwidth]{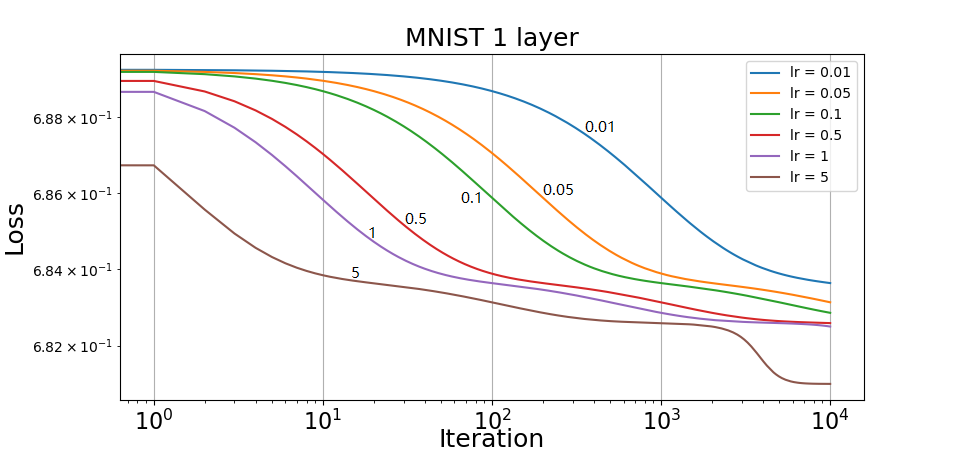}%
\end{minipage}%
\begin{minipage}[h]{0.5\linewidth}
\centering
\includegraphics[width=1\columnwidth]{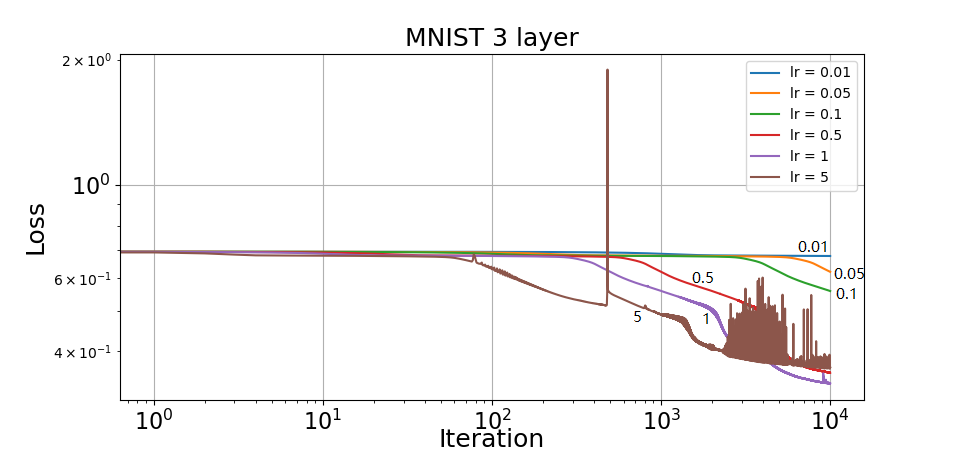}%
\end{minipage}

\caption{Dependence of Stable Learning Rates on Network Structure. {\bf [Left]}: Gradient descent experiment on the 1-layer CNN (see text for details). All learning rates are stable. {\bf [Right]}: Gradient descent experiment on the 3-layer CNN. Restrained instability occur at learning rate = 5, which is shown with the spikes and oscillations in the loss.}
\label{fig:mnist_layer_plot}
\end{figure}

{\bf Dependence of Stable Learning Rates on Layers/Network Structure}: As shown in Section~\ref{subsec:multilayer_restrained}, restrained instabilities in multi-layer networks arise locally at a single layer and through convolution can be spread to multiple layers/kernels. Therefore, restrained instabilities have a greater chance to arise and propagate in larger multi-layer networks, even at learning rates for which smaller networks can be stable. We now give empirical validation of this observation. We conduct experiments on 1-layer and 3-layer VGG-style CNN that is trained on the MNIST dataset (select data with labels 0 and 1). There is only 1 channel in each layer. The kernel size is 3x3 and we use the Swish activation with $\beta = 1$ here.  We replicate the cross entropy experiment in  \citep{DBLP:journals/corr/abs-2103-00065} on page 5 in Figure 5 under the same settings. As in \citep{DBLP:journals/corr/abs-2103-00065}, we optimize using full batch gradient descent (no stochasticity) without regularization (weight decay). In Figure~\ref{fig:mnist_layer_plot}, we plot the loss versus iteration at various learning rates for both the 1-layer and 3-layer networks. We observe that the upper bound for stable learning becomes lower with the increase in parameters. The 1-layer network is stable for all the chosen learning rates, however, the 3-layer network is unstable with restrained instabilities for a learning rate of $5$, even though the 1-layer network is stable at this learning rate.

\cut{
More complex networks have a greater chance for restrained instabilities to spread even at learning rates for the less complex network is stable. 
}

\begin{figure}[h]
\begin{center}
\includegraphics[width=\textwidth]{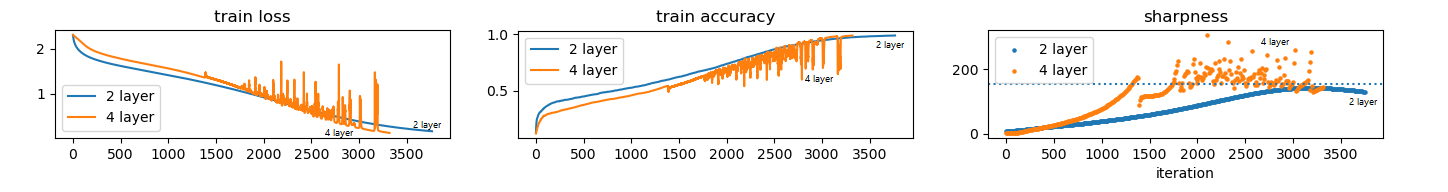}
\end{center}
\caption{Apparent Stability Without Regularization. The blue curves are for the stable learning rate found in the \citep{DBLP:journals/corr/abs-2103-00065} for a 2-layer FC network trained on CIFAR-5k without weight decay, suggesting stability can be achieved without weight decay. However, if one trains a more complex network (here doubling the layers), there are more chances for restrained instabilities to arise, and the stable learning rate now becomes unstable (in orange).}
\label{fig:eigen}
\end{figure}

{\bf Is Regularization Needed for Stability?} We discuss the role of regularization in stability, a prediction by our theory not known in the literature on EoS. From the stability condition in \eqref{eq:alpha_bounds}, we draw the conclusion that weight decay is necessary for the training process to be stable. But in the cross-entropy experiment in Figure 5 on page 5 in \citep{DBLP:journals/corr/abs-2103-00065}, it appears that the training process could be stable without regularization for a small enough learning rate. Our explanation for this apparent divergence is that the network structure is so simple that as soon as the instability kicks the kernel out of the transition region, it does not have a chance to move back from the activated or non-activated, hence the oscillations do not arise.  When the network becomes more complex, there are more interconnections between kernels/weights and more possibilities for the data to kick the kernels back into the transitioning region, and for the restrained instabilities to propagate to more locations in the network. To confirm our theory, we replicate the experiment (cross-entropy experiment in Figure 5 on page 5 in \citep{DBLP:journals/corr/abs-2103-00065}) in which a stable learning rate in a real network is found without regularization. The learning rate is $2/150$, the network utilized is a two-layer fully-connected (FC) network with tanh activation, with 200 nodes in each layer, and the dataset is CIFAR-5k (a subset of CIFAR-10 \citep{cifar10} with only 5000 training images so as to fit on a standard GPU). We optimize using full batch gradient descent (no stochasticity) without regularization (weight decay). Now, we double the number of fully connected layers and conduct the experiment under the same settings with this new network. In Figure~\ref{fig:eigen}, we observe spikes in the loss and sharpness exceeding the stable bound indicating restrained instabilities. This could be because the more complicated network gives more chances for the instability to propagate to more of the network, causing kernels to go in and out of the unstable transitioning region multiple times, without adequate regularization/weight decay.

\begin{figure}[h]
\begin{minipage}[h]{0.5\linewidth}
\centering
\includegraphics[width=1\columnwidth]{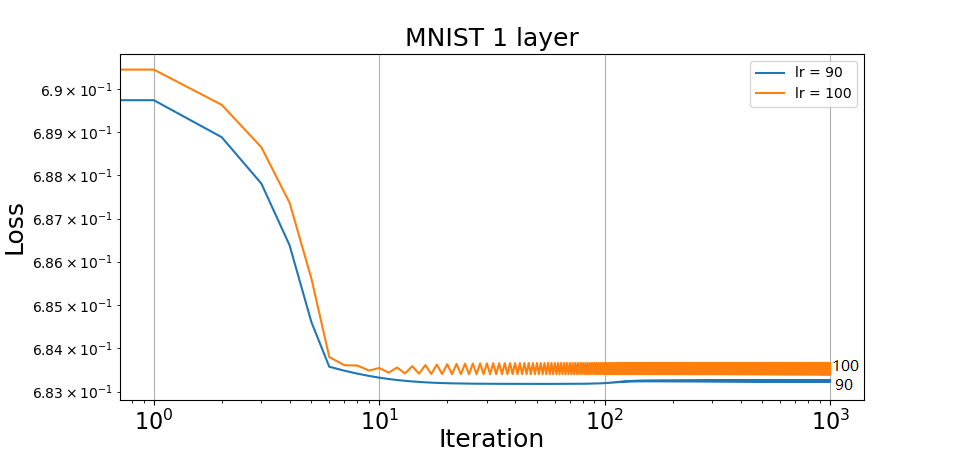}%
\end{minipage}%
\begin{minipage}[h]{0.5\linewidth}
\centering
\includegraphics[width=1\columnwidth]{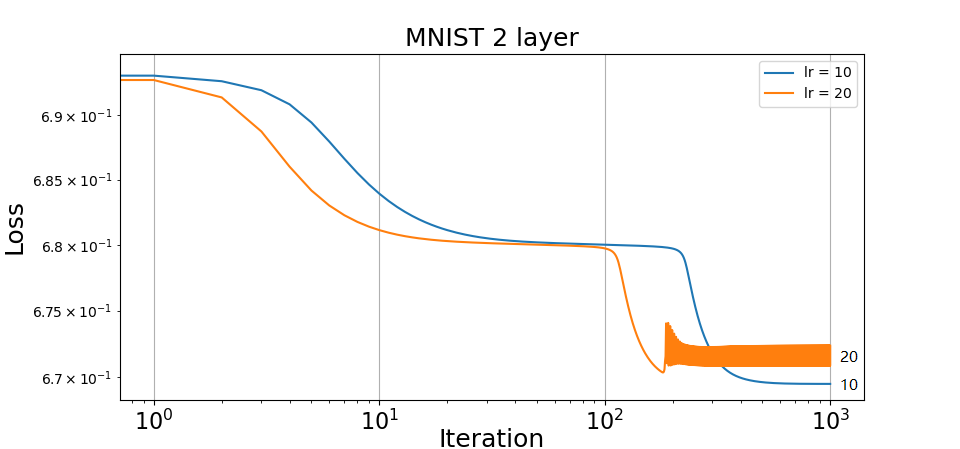}%
\end{minipage}

\caption{{Stability Bound Varies With Number of Layers. \bf [Left]}: Gradient descent with weight decay experiment on 1 layer CNN. The restrained stability boundary is between 90 and 100. {\bf [Right]}: Gradient descent with weight decay experiment on two-layer CNN. The restrained stability boundary is between 10 and 20.}
\label{fig:mnist_layer_w_plot}
\end{figure}

{\bf Stability Bounds Varies With Number of Layers (Regularized Case)}: As can be seen by our theory in Section~\ref{subsec:multilayer_restrained} (discussion following Theorem~\ref{thrm:mulilayer_gradient}), the bounds for regularization and learning rates change as the number of layers changes. In particular, the bounds in \eqref{eq:alpha_bounds} change by replacing $a$ with $\hat a = a\gamma$ where $\gamma$ is related to the gradient of the sub-network up to the layer of interest. We verify this empirically by determining the stability bounds on the learning rate empirically with regularization on a 1- and 2-layer CNN. We still use the selected MNIST dataset (mentioned in the first experiment of this sub-section) here. The kernel size is 3x3 and we use the Swish activation function. We choose $\beta = 1$ and $\alpha = 5e-4$. Results are shown in Figure~\ref{fig:mnist_layer_w_plot}, and confirm that the stability bounds change with the number of layers: in the 1-layer case, the stable learning rate is between 90 and 100, whereas for the two-layer network, it is more stringent and is between 10 and 20.

\section{Conclusion}
We discovered restrained numerical instabilities in standard training procedures of deep networks. In particular, we showed that epsilon-small errors inherent in floating point arithmetic can amplify and lead to divergence in final test accuracies, comparable to stochastic batch selection. Given such variations are comparable to accuracy gains reported in many papers, such instabilities could inflate or even suppress such gains, and further investigation is needed. Using a simplified multi-layer CNN models, which are amenable to linear analysis, we employed PDEs to explain how these instabilities might arise. We derived CFL conditions, which imposed conditions on the learning rate and weight decay. We showed that the rather than a full global blowup when CFL conditions are violated, the instabilities are \emph{restrained} and this is a result of properties of the non-linear gradient descent PDE. We showed that restrained instabilities are at the edge of stability, and thus our theory provided insights into that phenomenon, and further made new predictions.

\cut{
Our theory also predicted the relationship between the learning rate and weight decay (regularization) in overly regularized standard deep networks, which was validated empirically.
}

This study provided a step in principally choosing learning rates for stability and may serve as the basis for designing new optimization methods. We showed that numerical PDEs have promised to shed insight. The analysis showed how restrained instabilities can arise even in CNNs and how the phenomenon is more likely to arise in larger networks. Although empirically we showed how restrained instabilities could be eliminated with globally small learning rates, such instabilities are localized to space-time regions of the tensor space, suggesting the need to develop adaptive learning rate / optimization schemes tailored to have small rates in these local regions to ensure stability, while maintaining speed.

\section*{Acknowledgements}
This research was supported in part by Army Research Labs (ARL) W911NF-22-1-0267, Raytheon Technologies Research Center, and by the Intelligence Advanced Research Projects Activity (IARPA) via Department of Interior/ Interior Business Center (DOI/IBC) contract number 140D0423C0075. The U.S. Government is authorized to reproduce and distribute reprints for Governmental purposes notwithstanding any copyright annotation thereon. Disclaimer: The views and conclusions contained herein are those of the authors and should not be interpreted as necessarily representing the official policies or endorsements, either expressed or implied, of IARPA, DOI/IBC, or the U.S. Government.



\appendix

\section{Extended Discussion}
{\bf What is the relation between Section 3 and Section 4?} Section 3 presents a theory on how finite precision errors can be attenuated or amplified based on satisfying or not the CFL condition (condition on the learning rate) through PDE tools.   Note the continuum limit of SGD as the learning result goes to zero is a PDE. As such and for ease of illustration, a simple PDE is examined to illustrate the key ideas of the PDE framework. Section 4 presents that such amplification of finite precision errors is present in deep network optimization, and shows that this likely results from not satisfying the CFL condition (learning rate is too high) - the last experiment of the section.

{\bf What is the relation between Section 4 and Section 5?} Section 5 presents a detailed theoretical study on how the amplification/attenuation of finite precision errors from Section 4 can arise by applying methodology from Section 3.  The study reduces down restrained instabilities to its most basic manifestation in a simple CNNs trained on just a single image and shows how they arise. In particular, when the learning rate is chosen in a specific range, restrained instabilities (oscillating between stable and unstable regimes) arise and finite precision errors accumulate. For small enough learning rates, the instabilities disappear and finite errors are attenuated. This matches the behavior of the large networks in Section 4, suggesting the theory applies.

\cut{
{\bf What is the significance of this work?} We have taken a step in developing a theory for principally choosing and designing learning rate schemes from the perspective of numerical stability, showing in particular, the relevance of numerical PDE theory, which has not been explored before. The study also suggests the exploration of adaptive optimization schemes that are constructed to ensure numerical stability. Whether eliminating restrained instabilities in training results in better performance (e.g., generalization, training time, robustness, reduced variance, etc) is an open question for future exploration. It may be well the case that such stabilities are helpful in training, but this requires further investigation and our theory provides a starting point for answering that question. The work also discovered the phenomena of restrained instabilities and error amplification, and significant test accuracy variance resulting from that phenomena. Whether eliminating the instability reduces stochastic test variance from batch selection / initialization is an open question. Moreover, given that there are several papers in the literature where performance reported is on the order of the variance reported in this paper, our result may challenge the significance of those results, and further investigation is needed.
}

\section{Empirical Validation: All Seeds in Experiments of Section 4 are Fixed}
\label{app:seeds_fixed}
We verify that all random seeds in SGD and the deep learning framework (e.g., initialization, batch selection, CUDA seeds, etc) have been fixed in the experiments in Section~\ref{sec:expts_deep_net_instability}.  This would indicate that the only source of variance across trials is due to the introduced floating point perturbation. To verify that, we run experiments for multiple trials when $k=1$. Test accuracy over epochs from each trial is shown in the following Table~\ref{tab:epoch}. This confirms that the dynamic trajectories of weights strictly coincide, indicating all seeds have been fixed.

\begin{table}[h]
	\centering  
	\caption{Accuracy at different epochs over different trials.  This indicates all seeds inducing randomness have been fixed.}   
	\label{tab:epoch}
	\footnotesize
	\begin{tabular}{|c|c|c|c|c|c|c|c|c|c|c|c|}  
		\hline  
		Epoch & 0 & 20 & 40 & 60 & 80 & 100 & 120 & 140 & 160 & 180 & 200\\  
		\hline
		Trial 1& 33.84 & 79.22 & 90.90  & 91.19 & 93.07 & 93.33 & 93.38 & 93.47 & 93.44 & 93.39 & 93.47 \\
		\hline
		Trial 2& 33.84 & 79.22 & 90.90  & 91.19 & 93.07 & 93.33 & 93.38 & 93.47 & 93.44 & 93.39 & 93.47 \\
		\hline
		Trial 3& 33.84 & 79.22 & 90.90  & 91.19 & 93.07 & 93.33 & 93.38 & 93.47 & 93.44 & 93.39 & 93.47 \\
		\hline
	\end{tabular}
\end{table}

To further strengthen this point, we provide additional results on $k' = k\times 2^n$ in Table \ref{tab:k}. Since base 2 (binary) is used in floating number, multiplying or dividing by the power of 2 will not introduce floating point arithmetic perturbation, therefore for any integer $k$, results on $k'$ and $k$ will have the same trajectories. This is another piece of evidence showing that the instability comes ONLY from floating point error. We also provided code in the supplementary (main4.py) for verification.
\begin{table}[h]
	\centering  
	\caption{Final Test Accuracy for different $k$. Notice $k$ that differ by a factor of a power of two have the same accuracy. This provides more evidence that all seeds have been fixed.}
	\label{tab:k}
	\footnotesize
	\begin{tabular}{|c|c|c|c|c|c|c|c|c|c|c|c|c|}  
		\hline  
		& k=1 & k=2 & k=3 & k=4 & k=5 & k=6 & k=7 & k=8 & k=9 & k=10 & k=11 & k=12\\  
		\hline
		Trial 1& 93.47 & 93.47 & 93.29  & 93.47 & 93.40 & 93.29 & 93.72 & 94.47 & 93.62 & 93.40 & 93.79 & 93.29 \\
		\hline
		Trial 2& 93.47 & 93.47 & 93.29  & 93.47 & 93.40 & 93.29 & 93.72 & 93.47 & 93.62 & 93.40 & 93.79 & 93.29 \\
		\hline
		Trial 3& 93.47 & 93.47 & 93.29  & 93.47 & 93.40 & 93.29 & 93.72 & 93.47 & 93.62 & 93.40 & 93.79 & 93.29 \\
		\hline
	\end{tabular}
\end{table}

\section{Supplementary Experimental Results for Section 4}
\label{app:extra_tables}
We provide supplementary experimental results for Section 4.  The bold titles correspond to the section in Section 4.2 for which the supplementary experiment belongs.

{\bf Section 4.2, Perturbed SGD with Swish}: Table~\ref{tab:var_swish_datasets} shows detailed results of the variability due to the floating point perturbation in comparison to batch selection with Swish activation for Section 4.2 {\bf Perturbed SGD with Swish}. As noted in Section 4.2, the Swish activation also results in significant variance due to floating point arithmetic.

\begin{table}[h]
\center
\tiny
\begin{sc}
\begin{tabular}{cccccccc}
	\toprule
    Seed & 1 & 2 & 3 & 4 & 5 & 6 & STD\\
    \midrule
     $k=1$ & 93.81 & 93.79 & 93.62 & 93.83 & 93.72 & 93.59 & 0.09 \\
	\midrule
	 $k=3$ & 93.52 & 93.94 & 93.29 & 93.59 & 93.73 & 93.60 & 0.20\\
	 \midrule
	 $k=5$ & 93.39 & 93.89 & 93.89 & 93.60 & 93.71 & 93.62 & 0.17\\
	 \midrule
	 $k=7$ & 93.35 & 93.59 & 93.52 & 93.48 & 93.37 & 93.43 & 0.08 \\
	 \midrule
	 $k=9$ & 93.49 & 93.50 & 93.70 & 93.79 & 93.55 & 93.71 & 0.12 \\
	 \midrule
	 $k=11$ & 93.84 & 93.71 & 93.72 & 93.91 & 93.71 & 93.56 & 0.11 \\
	 \midrule
	 STD & 0.19 & 0.15 & 0.18 & 0.14 & 0.13 & 0.082 & \\
	\bottomrule
\end{tabular}
\begin{tabular}{ccc}
	\toprule
    & Floating Pt & SGD \\ \midrule
    Vgg16BN FMNIST & $94.81\pm 0.09$ &  0.10  \\ \midrule
    Vgg16BN CIFAR-10 & $93.76\pm 0.13$ & 0.09 \\ \midrule
    ResNet56 FMNIST & $94.81\pm 0.12$ &  0.10\\ \midrule
    ResNet56 CIFAR-10 & $93.72\pm 0.15$ & 0.09\\
    \bottomrule
\end{tabular}
\end{sc}
\caption{{\bf [Left]}:Test accuracy variance over different seeds (batch selections) and different floating point perturbations (rows) for Resnet56 using Swish activation that is trained on CIFAR-10. STD is the standard derivation. $k$ indicates different version of perturbed SGD that each introduces a perturbation at the last significant bit of the gradient.{\bf [Right]}:A summary of results of variation of test accuracy for different floating point error in comparison to stochastic noise for different architectures and datasets.}
\label{tab:var_swish_datasets}
\end{table}

{\bf Section 4.2, Other Architectures/Datasets}: Table \ref{tab:var_swish_datasets} verifies that the variability due to the floating point perturbation generally exists across different network architectures and datasets. Here we repeated the same experiment for a different network (VGG16) and a different dataset (Fashion-MNIST). The Swish activation is used. Six different floating point noises are used and six different seeds are used as in Table \ref{tab:var_swish_datasets}. Average standard deviations for test accuracy variation for both the floating point perturbations and batch selection. The test accuracy variance for floating point noise is greater or similar to the stochastic variation due to batch selection. Note the average test accuracy and the average standard deviation over floating point perturbations over $k$ are reported. The variance of SGD due to batch selection is also reported.

{\bf Section 4.2, Adam optimizer}: 
Table \ref{tab:var_adam} verifies that the variability due to the floating point perturbation also exists in adaptive learning rate optimizers. Here we repeat the same experiment with Adam optimizer.  

\begin{table}[h]
\center
\tiny
\begin{tabular}{cccccccc}
	\toprule
    Seed & 1 & 2 & 3 & 4 & 5 & 6 & STD\\
    \midrule
     $k=1$ & 88.16 & 88.36 & 88.88 & 88.43 & 88.63 & 88.39 & 0.22 \\
	\midrule
	 $k=3$ & 88.68 & 88.52 & 88.45 & 88.32 & 88.56 & 87.81 & 0.28\\
	 \midrule
	 $k=5$ & 88.59 & 88.30 & 88.33 & 88.25 & 88.62 & 88.79 & 0.19\\
	 \midrule
	 $k=7$ & 88.71 & 88.68 & 88.74 & 88.11 & 88.32 & 88.78 & 0.25\\
	 \midrule
	 $k=9$ & 88.79 & 88.65 & 88.72 & 88.59 & 88.48 & 88.73 & 0.10 \\
	 \midrule
	 $k=11$ & 88.00 & 88.73 & 88.78 & 88.71 & 88.21 & 88.42 & 0.29 \\
	 \midrule
	 STD & 0.34 & 0.21 & 0.16 & 0.36 & 0.21 & 0.13 & \\
	\bottomrule
\end{tabular}
\caption{Test accuracy variance over different seeds (batch selections) and different floating point perturbations (rows) for Resnet56 using Swish activation that is trained on CIFAR-10. STD is the standard derivation with Adam optimzier.}
\label{tab:var_adam}
\end{table}

{\bf Section 4.2, 64 bit floating point representation}: 
Table \ref{tab:var_64bit} evaluates the variability due to the floating point perturbation on a data type with higher numerical precision. Here we repeat the same experiment float64 data type.  Results indicate that errors are not mitigated with increasing precision, suggesting an instability.

\begin{table}[h]
\center
\tiny
\begin{tabular}{cccccccc}
	\toprule
    Seed & 1 & 2 & 3 & 4 & 5 & 6 & STD\\
    \midrule
     $k=1$ & 93.10 & 93.63 & 93.31 & 93.45 & 93.26 & 93.60 & 0.19\\
	\midrule
	 $k=3$ & 93.35 & 93.16 & 93.42 & 93.43 & 93.20 & 93.49 & 0.12\\
	 \midrule
	 $k=5$ & 93.32 & 93.38 & 93.23 & 92.74 & 93.37 & 93.29 & 0.22\\
	 \midrule
	 $k=7$ & 93.05 & 93.39 & 93.46 & 93.20 & 93.58 & 93.26 & 0.17\\
	 \midrule
	 $k=9$ & 93.21 & 93.60 & 93.28 & 93.53 & 93.39 & 93.66 & 0.16 \\
	 \midrule
	 $k=11$ & 93.22 & 93.26 & 93.33 & 93.35 & 93.56 & 93.62 & 0.14 \\
	 \midrule
	 STD & 0.10 & 0.16 & 0.07 & 0.26 & 0.14 & 0.15 & \\
	\bottomrule
\end{tabular}
\caption{Test accuracy variance over different seeds (batch selections) and different floating point perturbations (rows) for Resnet56 using ReLU activation that is trained on CIFAR-10 with 64 bit precision using SGD.}
\label{tab:var_64bit}
\end{table}

\begin{figure}[h]
\centering
\includegraphics[width=0.43\columnwidth]{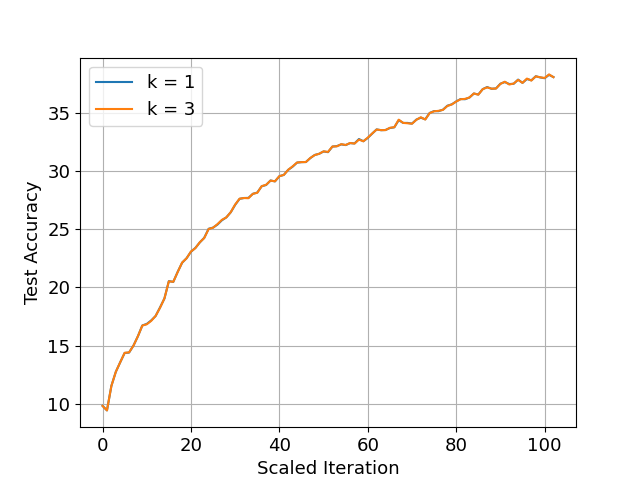}
\caption{Test accuracy plot for the stable learning rate 3.125e-6.  Shows that the optimization is not stuck in a local minimum, and the error amplification is eliminated because of the choice of learning rate that satisfies the CFL condition. Note the curves overlap.}
\label{fig:test_accuracy_stable}
\end{figure}

{\bf Section 4.2, Evidence of Divergence Due to Instability}: We verify that the optimization for the stable learning rate (3.125e-6) does not cause fluctuation for various $k$ in perturbed SGD due to being stuck at a local minimum.  To this end, we show the test accuracy plot for $k=1,3$ in Figure~\ref{fig:test_accuracy_stable} corresponding to the experiment on the right side of Figure 2 in the main paper. Note that the accuracy is increasing, indicating the optimization is not stuck at a local minimum.

\cut{
Although evident from Figure 2 (right) in the main paper, we also note from Figure~\ref{fig:test_accuracy_stable} that both accuracy curves for $k=1$ and $k=3$ overlap as the floating point errors are attenuated at this stable learning rate.  In contrast, if we display (see Figure~\ref{fig:test_accuracy}) the accuracy curves for the next highest learning rate (3.125e-5), the curves do not overlap, indicating that the floating point errors are amplified.

\cut{
To verify that the variability due to batch selection will disappear with stable learning rate, we plot the test accuracy for stable learning rate in the Fig\ref{fig:test_accuracy} left and test accuracy for unstable learning rate in the Fig\ref{fig:test_accuracy} right. The test accuracy for SGD($k=1$) and modified SGD($k=3$) are overlapped with each other for stable learning rate. In contrast, there is a gap between the test accuracy for SGD($k=1$) and modified SGD($k=3$) with unstable learning rate.
}

\begin{figure}[h]
\centering
\includegraphics[width=0.7\columnwidth]{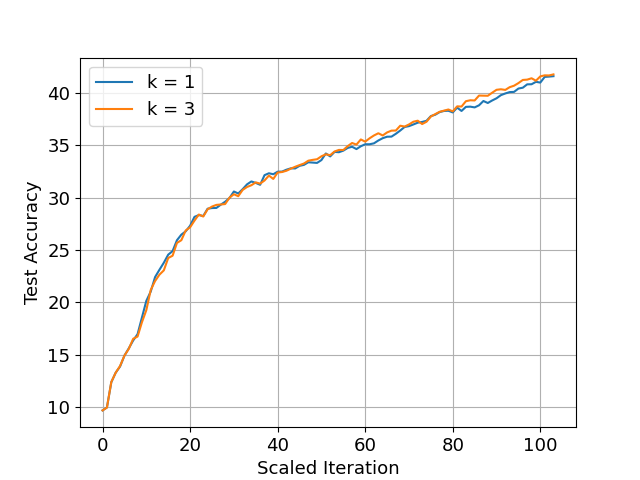}
\caption{Test accuracy plot for the unstable learning rate 3.125e-5, which shows that the curves do not overlap and thus floating point errors are amplified as the learning rate is not under the CFL condition. }
\label{fig:test_accuracy}
\end{figure}
}

\section{Derivation for Gradient Descent PDE and Stability Analysis}
\label{app:derivations}
\subsection{Derivation of Gradient PDE (\bf{Theorem 5.1})}
\label{sec:derivation_nonlinear}
Let $I : \R^2 \to \R$ be an input image to the network, $K : \R^2 \to \R$ be a convolution kernel (we assume $I$ and $K$ to be of finite support), $r: \R\to\R$ be the Swish activation, and $s : \R\to\R$ is the sigmoid function. We consider the following CNN, which a simple VGG:
\begin{equation}
f(I) = s\left[ \int_{\R^2} r(K \ast I)(x) \ud x \right].
\end{equation}

We consider the following regularized cross-entropy loss:
\begin{equation}
  L(K) = \ell( y, \hat y) + \frac 1 2 \alpha \|K\|^2_{\mathbb{L}^2},\quad
 \mbox{where} \quad \hat y = f(I)
 \label{eq:loss_app}
\end{equation}
where $\ell$ denotes the cross-entropy, the second term denotes the regularization ($\mathbb L^2$ norm squared of the kernel, $K$), and $\alpha>0$ and the second term is weight regularization. Define 
\begin{equation}
    g(I) = \int_{\R^2} r(K \ast I)(x) \ud x.
\end{equation}
To compute the gradient of the above, which is a functional, we employ methods from the calculus of variations \citep{https://doi.org/10.1002/zamm.19970770421}. We start by computing the variation of $g$ with respect to $K$; to denote the explicit dependence of $g$ on $K$, we write $g(I;K)$. The variation is defined as
\begin{equation}
    \delta g(I)\cdot \delta K  = \left.
    \der{g(I;K+\epsilon \delta K)}{\epsilon}
    \right|_{\epsilon=0},
\end{equation}
where $\delta K$ is a perturbation of $K$, i.e., in the tangent space of kernels at $K$.

Computing the variation of $g(I)$ with respect to $\delta K$ yields
\begin{align}
  \delta g(I) \cdot \delta K 
&=  \left.
    \der{\int_{\R^2} r((K+\epsilon \delta K)) \ast I)(x) \ud x}{\epsilon}
    \right|_{\epsilon=0}  \\
    &= \left.
    \der{\int_{\R^2} r(K\ast I + \epsilon \delta K\ast I)(x) \ud x}{\epsilon}
    \right|_{\epsilon=0} \\
    &= \left.\int r'(K\ast I + \epsilon \delta K\ast I)(x) \cdot \delta K \ast I(x) \ud x \right|_{\epsilon=0} \\   
  &= \int r'(K\ast I)(x) \cdot \delta K \ast I(x) \ud x \\
   &= \int_x \int_y r'(K\ast I)(x) I(x+z) \delta K (z) dz dx \\
   &= \int_z \delta K(z) \int_x r'(K\ast I)(x) I(x+z) \ud x \ud z \\
   &= \int_z \delta K(z) r'(K \ast I) \ast I(z) \ud z.
\end{align}
Note the derivative can be moved into the integrand because of the bounded integrand over all $\epsilon$. This expression allows us to see the gradient, $\nabla_K g(I)$ which is defined through the relation:
\begin{equation}
    \delta g(I)\cdot \delta K = 
    \int_z \nabla_K g(I)(z) \cdot \delta K(z) \ud z.
\end{equation}
Therefore,
\begin{equation}
\nabla_K g(I) = r'(K \ast I) \ast I.
\end{equation}
Therefore, using the chain rule, we have
\begin{equation}
  \nabla_K L(K) = \pder{\ell}{\hat y}(y,\hat y)  s'(g(I))  r'(K \ast I) \ast I + \alpha K.
\end{equation}
Note $s(x) = \frac{1}{1+e^{-(x-b)}}$ is the sigmoid (with bias $b$), and thus $s'(x) = \frac{e^{-(x-b)}}{[1+e^{-(x-b)}]^2} = s(x)[1-s(x)]$. Therefore, $s'(g(I)) = f(I)[1-f(I)] = \hat y(1-\hat y)$. Therefore, substituting the previous expression into the previous equation, we have
\begin{equation}
  \nabla_K L(K) = \hat y (1-\hat y)\pder{\ell}{\hat y}(y,\hat y)  r'(K \ast I) \ast I + \alpha K.
\end{equation}
For cross-entropy, we have $ \pder{\ell}{\hat y} = -\frac{y}{\hat y} + \frac{1-y}{1-\hat y} $ and thus $\hat y (1-\hat y)\pder{\ell}{\hat y} = \hat y - y$.
Therefore, substituting into the above equation, we have
\begin{equation}
 \nabla_K L(K) = (\hat y-y)r'(K \ast I) \ast I + \alpha K
\end{equation}

where $\hat y = f(I)$. The gradient descent PDE with respect to the loss \eqref{eq:loss_app} is 
    \begin{equation}
    \label{eq:PDE_no_constraint_app}
     \partial_t K = -\nabla_K L(K) = -(\hat y-y)r'(K \ast I) \ast I - \alpha K,
    \end{equation}
    where $r'$ denotes the derivative of the activation, and $t$ parametrizes the evolution. If we maintain the constraint that $K$ is finite support so that $K$ is zero outside $[-w/2,w/2]^2$ then the constrained gradient descent is 
    \begin{equation} \label{eq:PDE_constraint_app}
      \partial_t K = \left[ -(\hat y-y)r'(K \ast I) \ast I - \alpha K\right] \cdot W,
    \end{equation}
    where $W$ is a windowing function ($1$ inside $[-w/2,w/2]^2$ and zero outside).

\subsection{Derivation of Linearization of the Gradient PDE (\bf{Theorem 5.2})}
\label{sec:derivative_linear}
We assume that $r$ is the Swish function and $\beta>0$ is the parameter of the Swish activation:
\begin{equation}
  r(x) = \frac{x}{1+e^{-\beta x}}.
\end{equation}
We can show that the first derivative and the second derivative of Swish function are:
\begin{align}
  r'(x) = \frac{1+(1+\beta x)e^{-\beta x}}{(1+e^{-\beta x})^2},  \quad
  r''(x) = \frac{\beta e^{-\beta x} [\beta x(1-e^{-\beta x}) + 2(1+e^{-\beta x})] }{ (1+e^{-\beta x})^3 }.
\end{align}

We assume $K\ast I$ is near zero, so that we could express $r'(x)$ using Taylor expansion:
\begin{equation}
  r'(x) \approx r'(0) + r''(0) x, \quad \mbox{where}\quad 
  r'(0) = \frac 1 2, \quad r''(0) = \frac 1 2 \beta.
\end{equation}
In this case,
\begin{equation}
  [r'(K \ast I) \ast I]W \approx \frac 1 2 (\bar I + \beta [(K\ast I) \ast I] ) W
\end{equation}
where $\bar I$ is the input sum. Assuming $\hat y - y =: a$ is approximately constant, the constrained PDE becomes
\begin{equation} \label{eq:linear_PDE_app}
  \partial_t K = [-\frac a 2 (\bar I + \beta [(K\ast I) \ast I] )] W - \alpha K.
\end{equation}
Note that $KW = K$ since the initial $K$ is assumed be contained in the support of $W$, and the evolution will not change that.

\subsection{Derivation of CFL Conditions for Linearized PDE (\bf{Theorem 5.3, 5.4})}
\label{sec:derivation_bound}

Consider the forward Euler scheme of \eqref{eq:linear_PDE_app}:
\begin{equation}
\label{eq:discrete_app}
  K^{n+1} - K^n = \left[ -\frac a 2 \Delta t ( \bar I + \beta [(K^n\ast I) \ast I] ) \right] W - \Delta t \alpha K^n.
\end{equation}
where $n$ denotes the iteration number, and $\Delta t$ denotes the step size (learning rate in SGD).

We now apply the DFT. Note this requires some approximation of the original PDE defined on a continuous non-compact domain to a discrete domain. Namely, we approximate the linear convolution with a circular convolution as is done in Von-Neumann analysis (this assumption is true for compactly supported functions where the support is not growing). To apply the DFT we first compute the DFT of $(K\ast I)\ast I$ assuming periodic correlations:
\begin{align}
DFT\{(K^n\ast I) \ast I\} &= DFT\{K^n\ast I\}^* \cdot DFT\{I\} \\
&= (DFT\{K^n\}^* \cdot DFT\{I\})^* \cdot DFT\{I\} \\
&= DFT\{K^n\} \cdot DFT\{I\}^* \cdot DFT\{I\} \\
&= \hat K^n |\hat I|^2,
\end{align} 
where $K^\ast$ represents complex conjugate. The DFT of the discretization (\ref{eq:discrete_app}) of the linearized PDE is
    \begin{equation} \label{eq:DFT_app}
    \hat K^{n+1} - \hat K^n  = -\frac a 2 \Delta t ( \bar I + \beta \hat K^n |\hat I|^2 ) \ast \prod_{i=1}^{2} sinc(\frac{\omega_i}{2}) - \Delta t \alpha \hat K^n,
    \end{equation}
    where $\hat K^n$ denotes the DFT of $K^n$, and $\sinc$ denotes the sinc function. In the case that $w\to \infty$ (the window support becomes large),
    the DFT of $K$ can be written in terms of the amplifier $A$ as
    \begin{equation} \label{eq:DFT_amplifier_app}
      \hat K^{n+1}(\omega)  = A(\omega) \hat K^n(\omega)
                                       -\frac a 2 \Delta t  \bar I,\,\, \mbox{ where }\,\,
      A(\omega) = 1 -\Delta t \left(\alpha + \frac 1 2 a\beta |\hat I(\omega)|^2\right).
    \end{equation}
To be stable, $|A| < 1$.  Provided that $\alpha$ is large enough (when $a<0$), stability can be achieved with the condition:
\begin{equation} \label{eq:boundary}
  \Delta t < \frac{2}{ \alpha + \frac 1 2 a\beta |\hat I(\omega)|^2 }; 
\end{equation}
with $\alpha > -\frac 1 2 a \beta \max_{\omega}|\hat I(\omega)|^2$ (when $a<0$).
There are two conditions it has to be satisfied to be stable. First, as $\Delta t$ is positive, the denominator of the right side should be positive. That yields the lower bound for $\alpha$
\begin{equation}
    \alpha > - \frac 1 2 a \beta \max_{\omega} |\hat I(\omega)|^2.
\end{equation}
Then $\Delta t$ has to satisfy the CFL condition \eqref{eq:boundary} so that the system could be stable, which yields the upper bound for $\alpha$:
\begin{equation}
\alpha < \frac{2}{\Delta t} - \frac 1 2 a \beta \min_{\omega} |\hat I(\omega)|^2
\end{equation}

\subsection{Non-constant Linearization of $a=\hat y - y$ and Stability Analysis (Section 5.2)}
\label{sec:linear_a}

We consider a non-constant linear model of $K$ for $a$, so we linearize $a = \hat y - y$ around $K=0$:
\begin{align*}
  a &= \hat y - y = s(g_K(I)) - y \\
  &\approx s(g_0(I)) + \ip{s'(g_0(I)) \nabla_K g_0(I)}{K}{} - y \\
  &= s(0) + \ip{s'(0) r'(0) \ast I}{K}{} - y \\
  &= s(0) - y + \frac 1 2 s'(0) \bar I \bar K,
\end{align*}
where the subscript on $g$ indicates the dependence on the given kernel.
The gradient descent PDE following from the linearization above then becomes:
\begin{equation} 
  \partial_t K = \left[-\frac 1 2 [ s(0) - y + \frac 1 2 s'(0) \bar I \bar K ]
(\bar I + \beta [(K\ast I) \ast I] ) - \alpha K \right] W.
\end{equation}
Considering only linear terms and ignoring the constant (w.r.t $K$) terms, this becomes:
\begin{equation} \label{eq:Kevol}
  \partial_t K = \left[-\frac 1 2 [s(0)-y] \beta  [(K\ast I) \ast I] - \frac 1 4
  s'(0)\bar I^2 \bar K - \alpha K \right] W.
\end{equation}

Using forward Euler and computing the Fourier transform yields:
\begin{equation}
  \hat K^{n+1} = A(\omega) \hat K^n(\omega),
\end{equation}
where
\begin{equation}
  A(\omega) = 1 - \Delta t 
  \begin{cases}
    \frac 1 2 |\hat I(0)|^2 \left( \beta[s(0)-y] + \frac 1 4 s'(0)\delta(\omega) \right) +
    \alpha & \omega = 0 \\
    \frac 1 2 [s(0)-y]\beta |\hat I(\omega)|^2 + \alpha & \omega \neq 0 
  \end{cases},
\end{equation}
where $\delta$ is a Dirac delta function.  Note $s(0)-y$ can be either
positive or negative, so similar to the previous case where $a=\pder{\ell}{\hat y}s'$ was assumed
constant, the process can be unstable without regularization.

We approximate $\delta$ with a sinc function (Fourier transform of a
rectangular pulse, which is the case if the constant is defined on
just finite support), i.e.,
\begin{equation}
  \delta(\omega) = \frac{1}{2\pi L^2}
  \sinc{ \left(\frac{\omega_1}{2\pi L} \right) }
  \sinc{ \left(\frac{\omega_2}{2\pi L} \right) }.
\end{equation}
We can write $A$ as
\begin{equation}
  A(\omega) = 1 - \Delta t 
  \left(
    \frac 1 2 [s(0)-y]\beta |\hat I(\omega)|^2 + \alpha +
    \frac 1 4 s'(0)|\hat I(0)|^2 \delta(\omega)
  \right).
\end{equation}
The sign of $\frac 1 2 [s(0)-y]\beta |\hat I(\omega)|^2 + \frac 1 4
s'(0)|\hat I(0)|^2 \delta(\omega)$ can be either positive or negative,
therefore, in the case of no regularization, the process can be unstable.

This analysis shows that treating $a=\hat y - y$ as a more general non-constant linear function leads to similar CFL conditions and conclusions as the constant case in the main paper.




\subsection{Linearization of the PDE in the ``Activated'' Regime ({\bf Section 5.4})}
\label{subsec:linear_activate}
Suppose that $K\ast I$ is positive and away from zero, then $r'(K\ast I)=1$. The non-linear PDE \eqref{eq:PDE_no_constraint_app} reduces to
\begin{equation}
    \partial_t K = -a\bar I  - \alpha K,
\end{equation}
where $a = \hat y - y$.  Linearizing this as in Theorem 4.2 gives $a=s(0)-y + 0.5s'(0) \bar K \bar I$, gives:
\begin{equation}
    \partial_t K = -(s(0)-y)\bar I - \frac 1 8  \bar I^2 \bar K - \alpha K.
\end{equation}
For stability analysis, we can ignore the constant with respect to $K$ term.  One can show that in the DFT domain, the update of $K_t$ is given by 
\[
\hat K^{n+1}(\omega) = A(\omega) \hat K^n(\omega),
\]
where the amplifier factor is 
\[
A(\omega) = 
\begin{cases}
    1- \Delta t (\alpha +  \bar I^2/8) & \omega = 0 \\
    1-\alpha \Delta t & \omega \neq 0
\end{cases}.
\]
For stability, $|A|<1$, which implies the following conditions:
\[
\alpha > -\frac{\bar I^2}{8} \quad \mbox{and}\quad 
\alpha < \min\left\{ \frac{2}{\Delta t} - \frac{\bar I^2}{8}, \frac{2}{\Delta t} \right\} = 
\frac{2}{\Delta t} - \frac{\bar I^2}{8}.
\]

\section{Stability Conditions for Nesterov GD and Experiments }
\label{sec:nesterov}

\subsection{Derivation of Stability Conditions for Nesterov GD (Theroem 5.6)}

We start with discretizing PDE with momentum (\ref{eq:pde_momentum}) using a fully explicit scheme. Specifically, we use central difference approximations for both time derivatives giving a second order discretization in time:
\begin{equation}
\frac{K(t + \Delta t) - 2K(t) + K(t - \Delta t)}{\Delta t^2} + d \frac{K(t + \Delta t) - K(t - \Delta t)}{2\Delta t} = - \nabla_K L[K(t)]
\end{equation}
which leads to the following update:
\begin{equation}\label{eq:pde_discretize}
K^{n+1} = K^{n} + \frac{2 - d\Delta t}{2 + d\Delta t}\Delta K^n - \frac{2\Delta t^2}{2 + d\Delta t}\nabla_K L(K^n),
\end{equation}
where $K^n = K(n\Delta t)$, and $\Delta K^n = K^n-K^{n-1}$.
To obtain an update which more
closely resembles the classic two-part Nesterov recursion, we use a semi-implicit discretization. That is, we replace the explicit discretization $\nabla_K L(K^n)$ with a ``predicted estimate'' $\widehat{\nabla_K L(K^n)}$. This estimate is obtained by applying the same discretization but evaluating $\nabla_K L$ at the partial update $V^n = K^n + \frac{2 - d\Delta t}{2 + d\Delta t}\Delta K^n$, which is a ``look-ahead'' location. Using this strategy yields this two-step update as specified in the theorem. For more details see \citep{benyamin2020accelerated}.

To perform the stability analysis, we study linearizations of the PDE by using a similar procedure as in the stability analysis for gradient descent PDE presented in Section 5.3. The linearization of $\nabla_K L(V^n)$ is the same as in the linearized PDE \eqref{eq:linear_PDE_app}, given by
\begin{equation} \label{eq:linear_nesterov}
    \nabla_V L(V^n) = \left[ 
    -\frac a 2 (\bar I + \beta [(V\ast I) \ast I] ) - \alpha V
    \right] W        ,
\end{equation}

We now analyze the stability of the linearized equation using Von-Neumann analysis. To derive the stability conditions, we compute the spatial Discrete Fourier Transform (DFT) of the semi-implicit scheme \eqref{eq:semi_implicit_scheme} and solve for the gradient amplifier:

\begin{thrm}
The DFT of the semi-implicit scheme given in  \eqref{eq:semi_implicit_scheme}-\eqref{eq:semi_implicit_scheme2} of the linearized momentum PDE is given by 
\begin{align} 
\label{eq:DFT_semi_implicit}
\hat V^n &= \hat K^n + \frac{2 - d\Delta t}{2 + d\Delta t}(\hat K^n - \hat K^{n-1})
 \\
\hat K^{n+1} &= \hat V^n - \frac{2\Delta t^2}{2 + d\Delta t}z(\omega)\hat V^n
\end{align}
where the gradient amplifier $z$ is defined as 
\begin{equation}
    z(\omega) = \alpha + \frac 1 2 a\beta |\hat I(\omega)|^2.
\end{equation}

\end{thrm}

In order for the overall combined update updates to be a stable process, the gradient amplifier has to satisfy the following condition (for detailed proof, see\citep{benyamin2020accelerated}):
\begin{equation}
  \label{eq:implicit_CFL}
  \Delta t < \frac{2}{\sqrt{3\left(\alpha + \frac 1 2 a\beta |\hat I(\omega)|^2\right)}}.
\end{equation}

There are two conditions it has to be satisfied to be stable. First, as $\Delta t$ is positive, the denominator of the right side should be positive. That yields the lower bound for $\alpha$
\begin{equation}
    \alpha >  - \frac 1 2 a \beta \max_{\omega} |\hat I(\omega)|^2.
\end{equation}
Then $\Delta t$ has to satisfy the CFL condition (\ref{eq:implicit_CFL}) so that the system could be stable, which yields the upper bound for $\alpha$:
\begin{equation}
\alpha <  \frac{4}{3\Delta t^2} - \frac 1 2 a \beta \min_{\omega} |\hat I(\omega)|^2
\end{equation}

\subsection{Experimental Validation}

\begin{figure}[h]
\floatsetup{heightadjust=all, valign=c}
\begin{floatrow}
\ffigbox[0.4\textwidth][]{%
    \includegraphics[width=1\columnwidth]{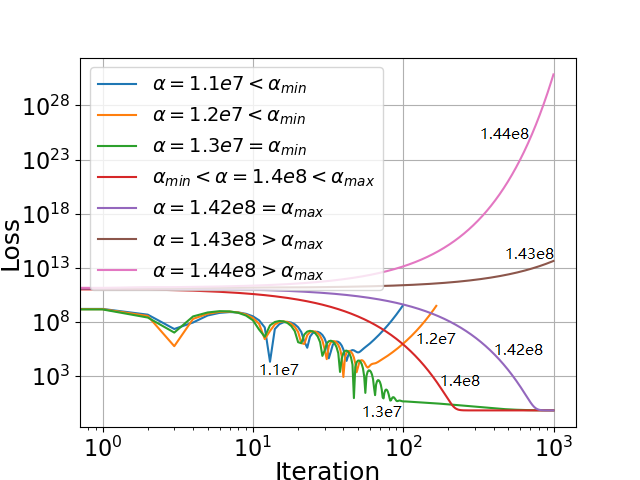}
}{%
  \captionof{figure}{Empirical validation of stability bounds for the linearized PDE \eqref{eq:linear_nesterov}.}%
  \label{fig:nesterov_PDE_stability_plot}%
}%
\capbtabbox[0.6\textwidth]{%
    \small
  	\begin{tabular}{ccccc}  
	\toprule
    Kernel & $\alpha_{min}^e $ & $\alpha_{min} $ & $\alpha_{max}^e$ &  $\alpha_{max}$\\
    \midrule
    16x16 & 3.9e6 & 1.07e9 & 1.42e8 & 1.33e8\\
	\midrule
	32x32 & 1.3e7 & 1.07e9 & 1.42e8 & 1.33e8\\
	\midrule
    64x64 & 3.9e7 & 1.07e9 & 1.42e8& 1.33e8\\
	\bottomrule
	\end{tabular}
}{%
  \captionof{table}{Comparison of the bounds on weight decay, $\alpha$, between the non-windowed linear PDE in \eqref{eq:nesterov_alpha_bounds} and the windowed linear PDE in \eqref{eq:linear_nesterov} found empirically.}%
  \label{tab:nesterov_alpha_window}
}
\end{floatrow}
\end{figure}

Figure~\ref{fig:nesterov_PDE_stability_plot} empirically validates the existence of upper and lower bounds on $\alpha$ for Nesterov momentum. We conduct a similar experiment as in section 5.2 to empirically validate this when the kernel is restricted to be finite support, i.e., the gradient is windowed. We empirically find the lower and upper bounds for $\alpha$ using the same method and compare it to the bounds in \eqref{eq:nesterov_alpha_bounds}.  We choose the $\Delta t = 1e-4, a = -0.5, \beta = 1$, and damping $d = \frac{2 - 2 \times 0.9}{\Delta t (1 + 0.9)}$ so that the momentum coefficient $\frac{2 - d\Delta t}{2 + d\Delta t} = 0.9$ is the common choice in deep learning training. Note that the choice of damping $d$ has no impact on the stability bounds. The input data is the same as the experiment in Section 5.2. See Table~\ref{tab:nesterov_alpha_window}, which verifies there are bounds on $\alpha$ for stability when using momentum as gradient descent.

\section{Multi-layer CNN Gradient Computation}
\label{app:multilayer}
We provide the derivation for the gradient with respect to any kernel $K_i$ in an arbitrary layer multi-layer convolutional neural network, and $K_i$ denotes the kernel in the $i^{\text{th}}$ layer. Our model is given as 
\begin{equation}
F(I) = F_{N}\circ F_{N-1}\circ \cdots \circ F_1(I),
\end{equation}
where $N$ is the number of layers, and
\begin{equation}
F_i(I_i) = r(K_i\ast I_i),\quad
I_i = F_{i-1} \circ \cdots \circ F_1(I).
\end{equation}
Define
\begin{equation}
F_{N,i+1}(I_{i+1}) = F_N\circ F_{N-1}\circ \cdots \circ F_{i+1}(I_{i+1}).
\end{equation}

We compute the variation of $F$ with respect to $K_i$. To do this, we note the following formulas, which are obtained through definitions of the gradient:
\begin{align}
    \delta F_{N,i+1}(I_{i+1}) \cdot \delta I_{i+1}(x) &= 
    \int \nabla_{I_{i+1}} F_{N,i+1}(x,y) \cdot \delta I_{i+1}(y) \ud y \\
    \delta I_{i+1}(y) = \delta F_i(I_i)\cdot \delta K_i(y) &=
    \int \nabla_{K_i} F_i(I_i)(y,z)\cdot \delta K_i(z) \ud z.
\end{align}
Combining the above formulas gives
\begin{equation}
\delta F \cdot \delta K_i = 
\int\left[ \int 
\nabla_{I_{i+1}} F_{N,i+1}(x,y) \nabla_{I_i} F_i(I_i)(y,z) 
\right] \ud y \cdot \delta K_i(z) \ud z.
\end{equation}
Using the previous formula and noting the definition of gradient, we see that
\begin{equation}
\nabla_{K_i} F(I)(x,z) = 
\int \nabla_{I_{i+1}} F_{N,i+1}(x,y) \nabla_{K_i} F_i(y,z) \ud y =: [\nabla_{I_{i+1}} F_{N,i+1} \cdot \nabla_{K_i} F_i] (x,z).
\end{equation}
Note
\begin{align}
    \delta F_i(I_i) \cdot \delta K_i(x) &= r'(K_i\ast I_i)(x) \delta K_i \ast I_i(x)
    = r'(K_i\ast I_i)(x) \int \delta K_i(y) I_i(x+y) \ud y;
\end{align}
therefore,
\begin{equation}
    \nabla_{K_i} F_i(I_i)(x,y) = r'(K_i\ast I_i)(x)I_i(x+y).
\end{equation}
Therefore,
\begin{align}
\nabla_{K_i} F(I)(x,z) = 
\int  \nabla_{I_{i+1}} F_{N,i+1}(x,y) r'(K_i\ast I_i)(y)I_i(y+z) \ud y.
\end{align}
Defining
\begin{equation}
    f(I) = s \left[ \int F(I)(x) \ud x \right],
\end{equation}
we get
\begin{align}
    \nabla_{K_i} f(I)(z) &= s' \int \nabla_{K_i} F(I)(x,z) \ud x \\
    &= s'\int \int \nabla_{I_{i+1}} F_{N,i+1}(x,y) r'(K_i\ast I_i)(y)I_i(y+z) \ud y \ud x \\
    &= s'\int \left[ \int \nabla_{I_{i+1}} F_{N,i+1}(x,y) \ud x \right]  r'(K_i\ast I_i)(y)I_i(y+z) \ud y \\
    &= s' [ \overline{\nabla_{I_{i+1}} F_{N,i+1}(I_{i+1})} \cdot r'(K_i\ast I_i) ] \ast I_i(z).
\end{align}
If the loss is defined as follows:
\begin{equation}
L = \ell( f(I), y ) + \alpha \sum_j \| K_j \|_{\mathbb L^2}^2,
\end{equation}
then,
\begin{equation}
    \nabla_{K_i} L = 
    (\hat y - y) \left[ \overline{\nabla_{I_{i+1}} F_{N,i+1}(I_{i+1})} \cdot r'(K_i\ast I_i) \right] \ast I_i + \alpha K_i,
\end{equation}
where
\begin{equation}
    \overline{\nabla_{I_{i+1}} F_{N,i+1}(I_{i+1})}(z) = \int \nabla_{I_{i+1}} F_{N,i+1}(I_{i+1})(x,z) \ud x.
\end{equation}
Note that $I_{i+1} = F_i(I_i) = r(K_i \ast I_i)$ depends on $K_i$, but $\nabla_{I_{i+1}}F_{N,i+1}$ does not depend on $K_i$. We show, for the purpose of stability analysis, we can approximate $\nabla_{I_{i+1}} F_{i+1}(I_{i+1})$ with a constant w.r.t $K_i$. Note
\begin{align}
    \delta F_{i+1}(I_{i+1}) \cdot \delta I_{i+1}(x) &= 
    r'(K_{i+1}\ast I_{i+1})(x)[K_{i+1} \ast \delta I_{i+1}](x) \\
    &= \int  r'(K_{i+1}\ast I_{i+1})(x) K_{i+1}(y)\delta I_{i+1}(x+y) \ud y \\
    &= \int r'(K_{i+1}\ast I_{i+1})(x) K_{i+1}(z-x)\delta I_{i+1}(z) \ud z.
\end{align}
Therefore,
\begin{equation}
    \nabla_{I_{i+1}} F_{i+1}(x,z) = r'(K_{i+1}\ast I_{i+1})(x) K_{i+1}(z-x).
\end{equation}
We consider the cases $K_i\ast I_i \approx 0$, $K_i\ast I_i \ll  0$, and $K_i\ast I_i \gg 0$. In these cases, $I_{i+1} \approx 0$, $I_{i+1} \approx 0$, and $I_{i+1} \approx K_i\ast I_i$, respectively. Then
\begin{equation}
     \nabla_{I_{i+1}} F_{i+1}(x,z) =  
     K_{i+1}(z-x)
        \begin{cases}
            \frac 1 2  & K_i\ast I_i \approx 0, K_i\ast I_i \ll 0 \\
            1 & K_i\ast I_i \gg 0
        \end{cases}.
\end{equation}
Therefore, the gradient at the $i+1$ layer approximately ceases dependence on $K_i$. We may apply a similar analysis to argue that $\nabla_{I_{i+k}} F_{i+k}(I_{i+k})$ is independent of $K_i$. Thus, the gradient $\nabla_{I_{i+1}} F_{N,i+1}(I_{i+1})$ is also independent of $K_i$.

For stability analysis, we thus treat $\nabla_{I_{i+1}} F_{N,i+1}(I_{i+1})$ independent of $K_i$ and we approximate it with a constant, $\gamma$, in space. Therefore, its enough to analyze stability of 
\begin{equation}
 \nabla_{K_i} L = 
   \hat a r'(K_i\ast I_i) \ast I_i + \alpha K_i,
\end{equation}
where $\hat a =  \gamma(\hat y - y)$. This follows the same stability analysis of the one-layer case; the result is that the stability conditions change by replacing $a$ with $\hat a$.

\vskip 0.2in
\bibliography{sample}

\begin{thebibliography}{61}
\providecommand{\natexlab}[1]{#1}
\providecommand{\url}[1]{\texttt{#1}}
\expandafter\ifx\csname urlstyle\endcsname\relax
  \providecommand{\doi}[1]{doi: #1}\else
  \providecommand{\doi}{doi: \begingroup \urlstyle{rm}\Url}\fi

\bibitem[Ahn et~al.(2022)Ahn, Zhang, and Sra]{https://doi.org/10.48550/arxiv.2204.01050}
Kwangjun Ahn, Jingzhao Zhang, and Suvrit Sra.
\newblock Understanding the unstable convergence of gradient descent, 2022.
\newblock URL \url{https://arxiv.org/abs/2204.01050}.

\bibitem[Arora et~al.(2022)Arora, Li, and Panigrahi]{arora2022understanding}
Sanjeev Arora, Zhiyuan Li, and Abhishek Panigrahi.
\newblock Understanding gradient descent on edge of stability in deep learning.
\newblock \emph{arXiv preprint arXiv:2205.09745}, 2022.

\bibitem[Bengio(2012)]{bengio2012practical}
Yoshua Bengio.
\newblock Practical recommendations for gradient-based training of deep architectures.
\newblock In \emph{Neural networks: Tricks of the trade}, pages 437--478. Springer, 2012.

\bibitem[Bengio(2015)]{bengio2015rmsprop}
Yoshua Bengio.
\newblock Rmsprop and equilibrated adaptive learning rates for nonconvex optimization.
\newblock \emph{corr abs/1502.04390}, 2015.

\bibitem[Benyamin et~al.(2020)Benyamin, Calder, Sundaramoorthi, and Yezzi]{benyamin2020accelerated}
Minas Benyamin, Jeff Calder, Ganesh Sundaramoorthi, and Anthony Yezzi.
\newblock Accelerated variational pdes for efficient solution of regularized inversion problems.
\newblock \emph{Journal of Mathematical Imaging and Vision}, 62\penalty0 (1):\penalty0 10--36, 2020.

\bibitem[Burger et~al.(2021)Burger, Ruthotto, Osher, et~al.]{burger2021connections}
M~Burger, L~Ruthotto, SJ~Osher, et~al.
\newblock Connections between deep learning and partial differential equations.
\newblock \emph{European Journal of Applied Mathematics}, 32\penalty0 (3):\penalty0 395--396, 2021.

\bibitem[Chaudhari and Soatto(2018)]{chaudhari2018stochastic}
Pratik Chaudhari and Stefano Soatto.
\newblock Stochastic gradient descent performs variational inference, converges to limit cycles for deep networks.
\newblock In \emph{2018 Information Theory and Applications Workshop (ITA)}, pages 1--10. IEEE, 2018.

\bibitem[Chaudhari et~al.(2018)Chaudhari, Oberman, Osher, Soatto, and Carlier]{chaudhari2018deep}
Pratik Chaudhari, Adam Oberman, Stanley Osher, Stefano Soatto, and Guillaume Carlier.
\newblock Deep relaxation: partial differential equations for optimizing deep neural networks.
\newblock \emph{Research in the Mathematical Sciences}, 5\penalty0 (3):\penalty0 1--30, 2018.

\bibitem[Chen and Bruna(2022)]{chen2022gradient}
Lei Chen and Joan Bruna.
\newblock On gradient descent convergence beyond the edge of stability.
\newblock \emph{arXiv preprint arXiv:2206.04172}, 2022.

\bibitem[Chen et~al.(2018)Chen, Rubanova, Bettencourt, and Duvenaud]{chen2018neural}
Ricky~TQ Chen, Yulia Rubanova, Jesse Bettencourt, and David Duvenaud.
\newblock Neural ordinary differential equations.
\newblock \emph{arXiv preprint arXiv:1806.07366}, 2018.

\bibitem[Chowdhery et~al.(2022)Chowdhery, Narang, Devlin, Bosma, Mishra, Roberts, Barham, Chung, Sutton, Gehrmann, et~al.]{chowdhery2022palm}
Aakanksha Chowdhery, Sharan Narang, Jacob Devlin, Maarten Bosma, Gaurav Mishra, Adam Roberts, Paul Barham, Hyung~Won Chung, Charles Sutton, Sebastian Gehrmann, et~al.
\newblock Palm: Scaling language modeling with pathways.
\newblock \emph{arXiv preprint arXiv:2204.02311}, 2022.

\bibitem[Cohen et~al.(2021)Cohen, Kaur, Li, Kolter, and Talwalkar]{DBLP:journals/corr/abs-2103-00065}
Jeremy~M. Cohen, Simran Kaur, Yuanzhi Li, J.~Zico Kolter, and Ameet Talwalkar.
\newblock Gradient descent on neural networks typically occurs at the edge of stability.
\newblock \emph{CoRR}, abs/2103.00065, 2021.
\newblock URL \url{https://arxiv.org/abs/2103.00065}.

\bibitem[Courant et~al.(1967)Courant, Friedrichs, and Lewy]{courant1967partial}
Richard Courant, Kurt Friedrichs, and Hans Lewy.
\newblock On the partial difference equations of mathematical physics.
\newblock \emph{IBM journal of Research and Development}, 11\penalty0 (2):\penalty0 215--234, 1967.

\bibitem[Courbariaux et~al.(2014)Courbariaux, Bengio, and David]{courbariaux2014training}
Matthieu Courbariaux, Yoshua Bengio, and Jean-Pierre David.
\newblock Training deep neural networks with low precision multiplications.
\newblock \emph{arXiv preprint arXiv:1412.7024}, 2014.

\bibitem[Damian et~al.(2022)Damian, Nichani, and Lee]{https://doi.org/10.48550/arxiv.2209.15594}
Alex Damian, Eshaan Nichani, and Jason~D. Lee.
\newblock Self-stabilization: The implicit bias of gradient descent at the edge of stability, 2022.
\newblock URL \url{https://arxiv.org/abs/2209.15594}.

\bibitem[Defazio et~al.(2014)Defazio, Bach, and Lacoste-Julien]{defazio2014saga}
Aaron Defazio, Francis Bach, and Simon Lacoste-Julien.
\newblock Saga: A fast incremental gradient method with support for non-strongly convex composite objectives.
\newblock In \emph{Advances in neural information processing systems}, pages 1646--1654, 2014.

\bibitem[Duchi et~al.(2011)Duchi, Hazan, and Singer]{duchi2011adaptive}
John Duchi, Elad Hazan, and Yoram Singer.
\newblock Adaptive subgradient methods for online learning and stochastic optimization.
\newblock \emph{Journal of machine learning research}, 12\penalty0 (Jul):\penalty0 2121--2159, 2011.

\bibitem[Goodfellow et~al.(2016)Goodfellow, Bengio, and Courville]{goodfellow2016deep}
Ian Goodfellow, Yoshua Bengio, and Aaron Courville.
\newblock \emph{Deep learning}.
\newblock MIT press, 2016.

\bibitem[Gupta et~al.(2015)Gupta, Agrawal, Gopalakrishnan, and Narayanan]{gupta2015deep}
Suyog Gupta, Ankur Agrawal, Kailash Gopalakrishnan, and Pritish Narayanan.
\newblock Deep learning with limited numerical precision.
\newblock In \emph{International conference on machine learning}, pages 1737--1746. PMLR, 2015.

\bibitem[Haber and Ruthotto(2017)]{haber2017stable}
Eldad Haber and Lars Ruthotto.
\newblock Stable architectures for deep neural networks.
\newblock \emph{Inverse problems}, 34\penalty0 (1):\penalty0 014004, 2017.

\bibitem[Han et~al.(2018)Han, Jentzen, and Weinan]{han2018solving}
Jiequn Han, Arnulf Jentzen, and E~Weinan.
\newblock Solving high-dimensional partial differential equations using deep learning.
\newblock \emph{Proceedings of the National Academy of Sciences}, 115\penalty0 (34):\penalty0 8505--8510, 2018.

\bibitem[He et~al.(2016)He, Zhang, Ren, and Sun]{He_2016_CVPR}
Kaiming He, Xiangyu Zhang, Shaoqing Ren, and Jian Sun.
\newblock Deep residual learning for image recognition.
\newblock In \emph{Proceedings of the IEEE Conference on Computer Vision and Pattern Recognition (CVPR)}, 2016.

\bibitem[Jastrzebski et~al.(2020)Jastrzebski, Szymczak, Fort, Arpit, Tabor, Cho, and Geras]{DBLP:journals/corr/abs-2002-09572}
Stanislaw Jastrzebski, Maciej Szymczak, Stanislav Fort, Devansh Arpit, Jacek Tabor, Kyunghyun Cho, and Krzysztof~J. Geras.
\newblock The break-even point on optimization trajectories of deep neural networks.
\newblock \emph{CoRR}, abs/2002.09572, 2020.
\newblock URL \url{https://arxiv.org/abs/2002.09572}.

\bibitem[Johnson and Zhang(2013)]{johnson2013accelerating}
Rie Johnson and Tong Zhang.
\newblock Accelerating stochastic gradient descent using predictive variance reduction.
\newblock \emph{Advances in neural information processing systems}, 26:\penalty0 315--323, 2013.

\bibitem[Karniadakis et~al.(2021)Karniadakis, Kevrekidis, Lu, Perdikaris, Wang, and Yang]{karniadakis2021physics}
George~Em Karniadakis, Ioannis~G Kevrekidis, Lu~Lu, Paris Perdikaris, Sifan Wang, and Liu Yang.
\newblock Physics-informed machine learning.
\newblock \emph{Nature Reviews Physics}, 3\penalty0 (6):\penalty0 422--440, 2021.

\bibitem[Khoo et~al.(2021)Khoo, Lu, and Ying]{khoo2021solving}
Yuehaw Khoo, Jianfeng Lu, and Lexing Ying.
\newblock Solving parametric pde problems with artificial neural networks.
\newblock \emph{European Journal of Applied Mathematics}, 32\penalty0 (3):\penalty0 421--435, 2021.

\bibitem[Kingma and Ba(2014)]{kingma2014adam}
Diederik~P Kingma and Jimmy Ba.
\newblock Adam: A method for stochastic optimization.
\newblock \emph{arXiv preprint arXiv:1412.6980}, 2014.

\bibitem[Krizhevsky et~al.()Krizhevsky, Nair, and Hinton]{cifar10}
Alex Krizhevsky, Vinod Nair, and Geoffrey Hinton.
\newblock Cifar-10 (canadian institute for advanced research).
\newblock URL \url{http://www.cs.toronto.edu/~kriz/cifar.html}.

\bibitem[Krizhevsky et~al.(2009)Krizhevsky, Hinton, et~al.]{krizhevsky2009learning}
Alex Krizhevsky, Geoffrey Hinton, et~al.
\newblock Learning multiple layers of features from tiny images.
\newblock 2009.

\bibitem[Lao et~al.(2020)Lao, Zhu, Wonka, and Sundaramoorthi]{lao2020channel}
Dong Lao, Peihao Zhu, Peter Wonka, and Ganesh Sundaramoorthi.
\newblock Channel-directed gradients for optimization of convolutional neural networks.
\newblock \emph{arXiv preprint arXiv:2008.10766}, 2020.

\bibitem[Latz(2021)]{latz2021analysis}
Jonas Latz.
\newblock Analysis of stochastic gradient descent in continuous time.
\newblock \emph{Statistics and Computing}, 31\penalty0 (4):\penalty0 1--25, 2021.

\bibitem[Lei et~al.(2017)Lei, Ju, Chen, and Jordan]{lei2017non}
Lihua Lei, Cheng Ju, Jianbo Chen, and Michael~I Jordan.
\newblock Non-convex finite-sum optimization via scsg methods.
\newblock \emph{arXiv preprint arXiv:1706.09156}, 2017.

\bibitem[Li et~al.(2022)Li, Wang, and Li]{https://doi.org/10.48550/arxiv.2207.12678}
Zhouzi Li, Zixuan Wang, and Jian Li.
\newblock Analyzing sharpness along gd trajectory: Progressive sharpening and edge of stability, 2022.
\newblock URL \url{https://arxiv.org/abs/2207.12678}.

\bibitem[Long et~al.(2018)Long, Lu, Ma, and Dong]{long2018pde}
Zichao Long, Yiping Lu, Xianzhong Ma, and Bin Dong.
\newblock Pde-net: Learning pdes from data.
\newblock In \emph{International Conference on Machine Learning}, pages 3208--3216. PMLR, 2018.

\bibitem[Luo et~al.(2019)Luo, Xiong, and Liu]{luo2018adaptive}
Liangchen Luo, Yuanhao Xiong, and Yan Liu.
\newblock Adaptive gradient methods with dynamic bound of learning rate.
\newblock In \emph{International Conference on Learning Representations}, 2019.
\newblock URL \url{https://openreview.net/forum?id=Bkg3g2R9FX}.

\bibitem[Ma et~al.(2022)Ma, Kunin, Wu, and Ying]{https://doi.org/10.48550/arxiv.2204.11326}
Chao Ma, Daniel Kunin, Lei Wu, and Lexing Ying.
\newblock Beyond the quadratic approximation: the multiscale structure of neural network loss landscapes, 2022.
\newblock URL \url{https://arxiv.org/abs/2204.11326}.

\bibitem[Naumann(1997)]{https://doi.org/10.1002/zamm.19970770421}
J.~Naumann.
\newblock Troutman, j. l.: Variational calculus and optimal control. optimization with elementary convexity. second edition. new york etc., springer-verlag 1996. xv, 461 pp., 87 figs., dm 84,00. isbn 0-387-94511-3 (undergraduate texts in mathematics).
\newblock \emph{ZAMM - Journal of Applied Mathematics and Mechanics / Zeitschrift für Angewandte Mathematik und Mechanik}, 77\penalty0 (4):\penalty0 316--316, 1997.
\newblock \doi{https://doi.org/10.1002/zamm.19970770421}.
\newblock URL \url{https://onlinelibrary.wiley.com/doi/abs/10.1002/zamm.19970770421}.

\bibitem[Osher et~al.(2018)Osher, Wang, Yin, Luo, Barekat, Pham, and Lin]{osher2018laplacian}
Stanley Osher, Bao Wang, Penghang Yin, Xiyang Luo, Farzin Barekat, Minh Pham, and Alex Lin.
\newblock Laplacian smoothing gradient descent.
\newblock \emph{arXiv preprint arXiv:1806.06317}, 2018.

\bibitem[Rakhlin et~al.(2011)Rakhlin, Shamir, and Sridharan]{rakhlin2011making}
Alexander Rakhlin, Ohad Shamir, and Karthik Sridharan.
\newblock Making gradient descent optimal for strongly convex stochastic optimization.
\newblock \emph{arXiv preprint arXiv:1109.5647}, 2011.

\bibitem[Ramachandran et~al.(2017)Ramachandran, Zoph, and Le]{ramachandran2017searching}
Prajit Ramachandran, Barret Zoph, and Quoc~V Le.
\newblock Searching for activation functions.
\newblock \emph{arXiv preprint arXiv:1710.05941}, 2017.

\bibitem[Richtmyer and Morton(1994)]{richtmyer1994difference}
Robert~D Richtmyer and Keith~W Morton.
\newblock Difference methods for initial-value problems.
\newblock \emph{Malabar}, 1994.

\bibitem[Rudin et~al.(1992)Rudin, Osher, and Fatemi]{rudin1992nonlinear}
Leonid~I Rudin, Stanley Osher, and Emad Fatemi.
\newblock Nonlinear total variation based noise removal algorithms.
\newblock \emph{Physica D: nonlinear phenomena}, 60\penalty0 (1-4):\penalty0 259--268, 1992.

\bibitem[Ruthotto and Haber(2020)]{ruthotto2020deep}
Lars Ruthotto and Eldad Haber.
\newblock Deep neural networks motivated by partial differential equations.
\newblock \emph{Journal of Mathematical Imaging and Vision}, 62\penalty0 (3):\penalty0 352--364, 2020.

\bibitem[Simonyan and Zisserman(2014)]{simonyan2014very}
Karen Simonyan and Andrew Zisserman.
\newblock Very deep convolutional networks for large-scale image recognition.
\newblock \emph{arXiv preprint arXiv:1409.1556}, 2014.

\bibitem[Sirignano and Spiliopoulos(2018)]{sirignano2018dgm}
Justin Sirignano and Konstantinos Spiliopoulos.
\newblock Dgm: A deep learning algorithm for solving partial differential equations.
\newblock \emph{Journal of computational physics}, 375:\penalty0 1339--1364, 2018.

\bibitem[Su et~al.(2014)Su, Boyd, and Candes]{su2014differential}
Weijie Su, Stephen Boyd, and Emmanuel Candes.
\newblock A differential equation for modeling nesterov’s accelerated gradient method: theory and insights.
\newblock \emph{Advances in neural information processing systems}, 27, 2014.

\bibitem[Sun et~al.(2021)Sun, Lao, Sundaramoorthi, and Yezzi]{sun2021accelerated}
Yuxin Sun, Dong Lao, Ganesh Sundaramoorthi, and Anthony Yezzi.
\newblock Accelerated pdes for construction and theoretical analysis of an sgd extension.
\newblock In \emph{The Symbiosis of Deep Learning and Differential Equations}, 2021.

\bibitem[Sun et~al.(2022)Sun, Lao, Sundaramoorthi, and Yezzi]{sun2022surprising}
Yuxin Sun, Dong Lao, Ganesh Sundaramoorthi, and Anthony Yezzi.
\newblock Surprising instabilities in training deep networks and a theoretical analysis.
\newblock In \emph{Advances in Neural Information Processing Systems}, 2022.
\newblock URL \url{https://openreview.net/pdf?id=Qi4vSM7sqZq}.

\bibitem[Sundaramoorthi and Yezzi(2018)]{sundaramoorthi2018variational}
Ganesh Sundaramoorthi and Anthony Yezzi.
\newblock Variational pdes for acceleration on manifolds and application to diffeomorphisms.
\newblock \emph{Advances in Neural Information Processing Systems}, 31, 2018.

\bibitem[Toulis et~al.(2016)Toulis, Tran, and Airoldi]{toulis2016towards}
Panos Toulis, Dustin Tran, and Edo Airoldi.
\newblock Towards stability and optimality in stochastic gradient descent.
\newblock In \emph{Artificial Intelligence and Statistics}, pages 1290--1298. PMLR, 2016.

\bibitem[Trefethen(1996)]{trefethen1996finite}
Lloyd~Nicholas Trefethen.
\newblock Finite difference and spectral methods for ordinary and partial differential equations.
\newblock 1996.

\bibitem[Weinan(2017)]{weinan2017proposal}
E~Weinan.
\newblock A proposal on machine learning via dynamical systems.
\newblock \emph{Communications in Mathematics and Statistics}, 5\penalty0 (1):\penalty0 1--11, 2017.

\bibitem[Weinan et~al.(2020)Weinan, Ma, Wojtowytsch, and Wu]{weinan2020towards}
E~Weinan, Chao Ma, Stephan Wojtowytsch, and Lei Wu.
\newblock Towards a mathematical understanding of neural network-based machine learning: what we know and what we don’t.
\newblock \emph{arXiv preprint arXiv:2009.10713}, 2020.

\bibitem[Wibisono et~al.(2016)Wibisono, Wilson, and Jordan]{wibisono2016variational}
Andre Wibisono, Ashia~C Wilson, and Michael~I Jordan.
\newblock A variational perspective on accelerated methods in optimization.
\newblock \emph{proceedings of the National Academy of Sciences}, 113\penalty0 (47):\penalty0 E7351--E7358, 2016.

\bibitem[Wilson et~al.(2016)Wilson, Recht, and Jordan]{wilson2016lyapunov}
Ashia~C Wilson, Benjamin Recht, and Michael~I Jordan.
\newblock A lyapunov analysis of momentum methods in optimization.
\newblock \emph{arXiv preprint arXiv:1611.02635}, 2016.

\bibitem[Wu et~al.(2018)Wu, Ma, and E]{NEURIPS2018_6651526b}
Lei Wu, Chao Ma, and Weinan E.
\newblock How sgd selects the global minima in over-parameterized learning: A dynamical stability perspective.
\newblock In S.~Bengio, H.~Wallach, H.~Larochelle, K.~Grauman, N.~Cesa-Bianchi, and R.~Garnett, editors, \emph{Advances in Neural Information Processing Systems}, volume~31. Curran Associates, Inc., 2018.
\newblock URL \url{https://proceedings.neurips.cc/paper_files/paper/2018/file/6651526b6fb8f29a00507de6a49ce30f-Paper.pdf}.

\bibitem[Xiao et~al.(2017)Xiao, Rasul, and Vollgraf]{xiao2017fashion}
Han Xiao, Kashif Rasul, and Roland Vollgraf.
\newblock Fashion-mnist: a novel image dataset for benchmarking machine learning algorithms.
\newblock \emph{arXiv preprint arXiv:1708.07747}, 2017.

\bibitem[Yang et~al.(2024)Yang, Sun, Sundaramoorthi, and Yezzi]{yang2024stabilizing}
Huizong Yang, Yuxin Sun, Ganesh Sundaramoorthi, and Anthony Yezzi.
\newblock Stabilizing the optimization of neural signed distance functions and finer shape representation.
\newblock \emph{Advances in Neural Information Processing Systems}, 36, 2024.

\bibitem[Zeiler(2012)]{zeiler2012adadelta}
Matthew~D Zeiler.
\newblock Adadelta: an adaptive learning rate method.
\newblock \emph{arXiv preprint arXiv:1212.5701}, 2012.

\bibitem[Zhang et~al.(2022)Zhang, Roller, Goyal, Artetxe, Chen, Chen, Dewan, Diab, Li, Lin, et~al.]{zhang2022opt}
Susan Zhang, Stephen Roller, Naman Goyal, Mikel Artetxe, Moya Chen, Shuohui Chen, Christopher Dewan, Mona Diab, Xian Li, Xi~Victoria Lin, et~al.
\newblock Opt: Open pre-trained transformer language models.
\newblock \emph{arXiv preprint arXiv:2205.01068}, 2022.

\bibitem[Zhu et~al.(2022)Zhu, Wang, Wang, Zhou, and Ge]{zhu2022understanding}
Xingyu Zhu, Zixuan Wang, Xiang Wang, Mo~Zhou, and Rong Ge.
\newblock Understanding edge-of-stability training dynamics with a minimalist example.
\newblock \emph{arXiv preprint arXiv:2210.03294}, 2022.

\end{thebibliography}

\end{document}